\documentclass{article}

\usepackage[final]{colm2026_conference}

\usepackage[T1]{fontenc}
\usepackage[utf8]{inputenc}

\usepackage{microtype}

\usepackage{graphicx}
\usepackage{hyperref}
\usepackage{url}

\usepackage{lineno}

\usepackage[table,xcdraw]{xcolor}
\usepackage{colortbl}

\definecolor{darkblue}{rgb}{0, 0, 0.5}
\hypersetup{colorlinks=true, citecolor=darkblue, linkcolor=darkblue, urlcolor=darkblue}

\definecolor{myred}{RGB}{220, 50, 47}
\definecolor{mygreen}{RGB}{133, 153, 0}
\definecolor{myblue}{RGB}{38, 139, 210}
\definecolor{myorange}{RGB}{203, 75, 22}

\usepackage{booktabs}
\usepackage{multirow}
\usepackage{multicol}
\usepackage{makecell}
\usepackage{longtable}

\usepackage{amsfonts}
\usepackage{amsmath}
\usepackage{nicefrac}

\usepackage{algorithm}
\usepackage{algpseudocode}
\usepackage{algorithmicx}

\usepackage{tcolorbox}
\usepackage{cleveref}
\usepackage{enumitem}
\usepackage{listings}
\usepackage{caption}

\makeatletter
\newcommand{\printfnsymbol}[1]{%
  \textsuperscript{\textcolor{black}{\@fnsymbol{#1}}}%
}
\makeatother

\newcommand{\spm}[1]{{\tiny$\pm$#1}}
\newcommand{\ssr}[1]{{\tiny$\pm$#1x}}

\newif\ifshowold   \showoldtrue     
\newif\iffinalmode \finalmodefalse 

\definecolor{crblue}{RGB}{0,64,186}
\definecolor{crgray}{gray}{0.48}

\title{Communication and Verification in LLM Agents towards \\Collaboration under Information Asymmetry}

\author{%
    Run Peng$^1$\thanks{Equal contributions.}\qquad Ziqiao Ma$^1$\printfnsymbol{1}\qquad Amy Pang$^1$ \qquad Sikai Li$^2$\thanks{Work done while at the University of Michigan.}  \\ 
    \textbf{Zhang Xi-Jia$^3$\printfnsymbol{2}\qquad Yingzhuo Yu$^4$\printfnsymbol{2}\qquad Cristian-Paul Bara$^{5,6}$\printfnsymbol{2}\qquad Joyce Chai$^1$} \\
    $^1$University of Michigan\;
    $^2$UNC, Chapel Hill\;
    $^3$Georgia Tech\;\\
    $^4$Apple\;
    $^5$Robert Bosch SRL\;
    $^6$Babe\c{s}-Bolyai University\;
}

\begin{document}

\ifcolmsubmission
\linenumbers
\fi

\maketitle

\begin{abstract}
Large language model (LLM) agents are increasingly used for goal-directed tasks, yet collaboration under asymmetric information remains underexplored. We adapt Einstein Puzzles into a tabletop game in which two agents hold complementary constraints and control different parts of the workspace. Solving each game requires agents to reason, communicate, and manipulate objects toward a shared goal. We fine-tune agents under four communicative action spaces and evaluate an environment-based reasoning verifier that filters candidate actions using physical affordances, communication history, and inferred constraints. In the strongest Llama 3.1 configuration, agents that can both provide and seek information achieve the highest success rate, while mismatched communication capabilities reduce performance. Verifier-assisted inference improves success rates across all evaluated action spaces. Some variants also achieve high success without communication, but their trajectories contain more rule-inconsistent actions and rejected placement attempts. In a human study with 30 participants, non-communicative agents require a similar number of steps as fully communicative agents but are rated as more confusing. These findings show that task success alone does not capture the quality of collaborative behavior and highlight the value of bidirectional communication and environment-derived verification. \url{https://github.com/Roihn/EinsteinPuzzles}
\end{abstract}
\section{Introduction}



In recent years, there has been growing interest in Large Language Model (LLM) agents (e.g., Web-Agents) and their diverse applications~\citep{wang2024survey}. While much of the current work focuses on action planning or goal completion in LLM agents~\citep{saycan2022arxiv,durante2024agent,song2023llm} (e.g., based on natural language instructions), their ability to collaborate with one another toward a shared goal remains underexplored. This paper addresses this gap by studying LLM agents in the context of task collaboration, particularly under conditions of information asymmetry.

Information asymmetry is a fundamental and pervasive feature of human interaction. In daily life, we often possess different  knowledge, perspectives, or intentions. We must reason about others’ knowledge and beliefs and coordinate our differences in order to collaborate~\citep{ma2023towards}. As LLM agents become increasingly integrated into real-world workflows — not just to perform tasks independently, but to act as human proxies and collaborators — it becomes essential to examine how well these agents can coordinate under asymmetric information, and what mechanisms might enhance their collaborative capabilities in such settings.




To this end, we adapt Einstein Puzzles~\citep{Groza2021}, a classical logic problem, into a tabletop environment where two agents must solve spatial and relational constraints, despite having partial, asymmetric information. Using a fine-tuning–plus–verifier framework, we equip LLM agents with different communicative abilities (e.g., asking, sharing, or both) and study their collaborative behaviors under various configurations.

Beyond communication design, we investigate whether the environment can provide useful verification signals during inference. Our verifiers check physical affordances, communication validity, and consistency with the agent’s known constraints. It is training-free, lightweight, and broadly applicable across various environments. We examine whether this simple yet generalizable approach can greatly improve collaboration among LLM agents.

Our empirical results show that the effects of communication depend on the model and reasoning configuration. In the strongest Llama 3.1 setting, Provide \& Seek achieves the highest success rate, while targeted seeking generally outperforms unprompted providing across matched settings. Mismatched communicative action spaces reduce collaboration performance. Verifier-assisted inference yields substantial gains and reduces rule-inconsistent and redundant actions. The human study further shows that similar task success and step counts can mask differences in interaction quality: participants report greater confusion with non-communicative agents than with fully communicative agents. Together, these findings motivate evaluating collaborative agents through both task outcomes and communication behavior.

\section{Related work}

\begin{figure*}[t!]
    \centering
    \includegraphics[width=\linewidth]{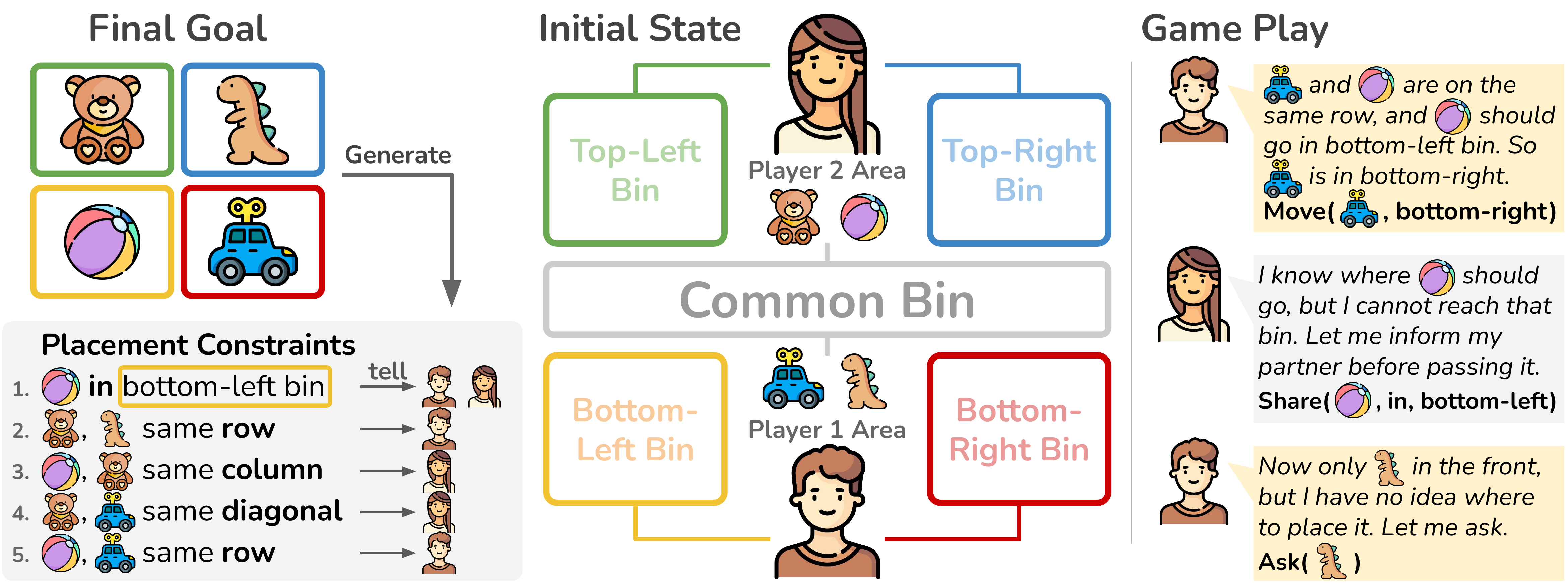}
    \caption{Illustration of our collaborative game. Each game features a final goal (top-left), where objects are assigned to specific goal bins (e.g. toy bear to top-left bin). Placement constraints are generated based on this final goal and are distributed to the two players. At the start, objects are randomly positioned in front of the players. Players must collaborate and communicate effectively to infer the final goal and accordingly complete the placement.}
    \label{fig:acl:game_intro}
    \vspace{-5pt}
\end{figure*}

\paragraph{Collaboration among LLM Agents}

The field of multi-agent coordination and communication has a long-established history~\citep{albrecht2018autonomous,gronauer2022multi,stone2000multiagent}, with applications increasingly extending to human-AI collaboration~\citep{carroll2019utility,puig2020watch}. Recently, multi-agent systems built upon Large Language Models (LLMs)~\citep{guo2024large,tran2025multi} have pushed the boundaries of collaborative intelligence in a wide range of downstream tasks, including collaborative coding~\citep{gao2024collabcoder,hong2024metagpt,qian2024chatdev}, social simulation~\citep{li2023camel,wu-etal-2024-shall,yang2024oasis,zhou2024sotopia}, and problem-solving~\citep{chen2023reconcile,qian2025scaling,wang2024sibyl,zhang2024building}.

However, most existing work focuses on settings with information transparency~\citep{chen2023reconcile,gao2024collabcoder,li2023camel,wu-etal-2024-shall,yang2024oasis}, one-way communication~\citep{qian2025scaling}, or information asymmetry within asymmetric role assignments~\citep{hong2024metagpt,qian2024chatdev,zhou2024sotopia}. In contrast, this work centers on LLM agents operating under information asymmetry in symmetric, collaborative roles, and includes a human study to assess real-world human-AI interaction. \citealp{liuautonomous} studied autonomous agents for collaborative tasks under information asymmetry, while there were no environments involved. As echoed in recent efforts~\citep{liuautonomous,zhou2024real}, addressing collaboration under information asymmetry—especially in human-AI contexts—remains a core challenge for LLM-based systems.

\paragraph{Test-time Compute in LLM}

Recent works have demonstrated the effectiveness of leveraging additional computational resources at inference time to improve response quality in LLMs. One line of research introduces a verification model to evaluate the correctness and utility of generated responses. Typically, this involves sampling multiple responses from an LLM and applying a best-of-$n$ strategy~\citep{charniak-johnson-2005-coarse,cobbe2021training}, where a verifier model selects the most appropriate response. Such approaches have been widely adopted across domains including mathematics~\citep{cobbe2021training,lifshitz2025multi,lightman2024lets,wang-etal-2024-math,yao2023tree}, code generation~\citep{mcaleese2024llm,wang2025training}, and web navigation~\citep{koh2024tree,putta2024agent}. We refer readers to~\citep{guan2024search,zhang2025and} for comprehensive reviews of verification-based methods.

While most existing approaches employ a separate value model, often another LLM, with~\citep{lifshitz2025multi,lightman2024lets,mcaleese2024llm,wang2025training,wang-etal-2024-math,zhang2024generative,zhou2310language} or without~\citep{yao2023tree,yu2023prompt} further training, we propose an alternative verification mechanism that directly leverages environmental feedback. In interactive tasks within simulated environments, the environment itself provides fine-grained, up-to-date, and objective signals about task progression and action validity. This enables a training-free, compute-efficient form of verification that naturally integrates with agent decision-making. 
\section{Collaboration Under Information Asymmetry}
\subsection{Understanding Einstein Puzzles on the tabletop setup}

Einstein Puzzles is a logical game requiring deductive reasoning and constraint satisfaction. We modify it into a collaborative game in which two agents work together to place a set of objects into designated bins. Each object has a target bin as its destination, but the information about these destinations is split between the two agents. Importantly, agents are unaware of which pieces of information their partner possesses. To succeed, agents must engage in rich communication and apply strategic reasoning to solve the task collaboratively.


The setup consists of a rectangular table with two collaborators sitting on opposite sides of the table (See Figure~\ref{fig:acl:game_intro}). The table is split into three regions: a region immediately in front of each collaborator and a common area in the center. Each collaborator can only reach the area directly in front of them and the common area in the center. They can also each reach the two corner bins on either side of the area in front of them, totaling four destination bins. In the beginning, several objects are distributed between the two players' bins, so they are exclusively reachable by one of the collaborators. No objects start in any bins.

We describe the task as a distributed constraint satisfaction problem. A goal configuration is defined, which is where the objects should be placed, and can be described with a set of constraints. There are four types of constraints that describe the relationship of a pair of objects or between an object and a bin. Given a pair of objects, it must either be:
 \begin{itemize} 
     \item \textbf{in the same bin}: both objects have the same destination.
     \item \textbf{in the same column}: the two objects must end up in bins on either exclusive side of the table but on the same side, left or right, from either collaborator's perspective.
     \item \textbf{in the same row}: both objects must end up on one of the collaborator's exclusive zones but in bins on opposite sides, left or right.
     \item \textbf{or on the same diagonal}: the two objects must end up in opposite areas and opposite sides.
 \end{itemize} 
 

The exhaustive set of rules between all pairs of objects is overly descriptive. We select a random, minimal subset of these rules (denoted as $\mathcal{C}$). A minimal subset $\mathcal{C}$ must satisfy the following:
\begin{itemize} 
    \item Eliminating any rule will describe more than one final configuration.
    \item Moving any object from the final configuration will violate at least one rule.
\end{itemize}
An additional starting constraint is given, which grounds one of the objects to a specific bin. Together, the minimal set of object pair rules and the one object-to-bin grounding uniquely describe the goal configuration. 

The set of constraints that uniquely determines the final placement of the objects is then divided between the two players (denoted as $\mathcal{C}_1$ and $\mathcal{C}_2$), creating \textbf{information asymmetry}. This distribution ensures that neither player can complete the task independently, but together, they possess all the necessary information to succeed, i.e., $\mathcal{C}_1 \cup \mathcal{C}_2 = \mathcal{C}$. 
Coupled with the fact that some objects must end on the opposite side of the table, this constitutes the disparity in knowledge and skills as described in \citep{bara-etal-2021-mindcraft}.

The two players take turns, each consisting of a single move. A move can be placing an object into an available destination bin or the center of the table, sharing a piece of information, or requesting information from their partner. Task performance is measured by the number of moves taken to complete the task. Since both object placement and information exchange count as a move, this scoring system encourages players to communicate concisely. The players’ objective is to achieve the final configuration using the least number of moves within the set of given constraints. 

\subsection{Information Providing \& Seeking}
We study two communicative actions in collaborative contexts with information asymmetry: information seeking and information providing. We want to investigate whether LLM agents are able to exchange necessary information with their partners, as well as perform collaborative reasoning for task completion. To systematically analyze the role of communication actions in this process, we design four action space configurations for communication as below:
\begin{enumerate} 
    \item \textbf{\textit{Information Providing Only:}} agents are allowed to proactively share their knowledge of constraints with their partners, but they cannot ask for this information. 
    \item \textbf{\textit{Information Seeking Only:}} agents can ask about constraints of specific objects with their partners. They are allowed to share a constraint only when their partner first asks about an object involved in it. In other words, agents cannot initiate the information providing.
    \item \textbf{\textit{Information Providing \& Seeking:}} agents can both share and ask freely.
    \item \textbf{\textit{No Information Exchange:}} no communicative actions are enabled. Agents can only choose to move an object or to pass the turn. Since neither player has complete knowledge about the placement constraints, this configuration inevitably involves random guessing for objects that may have goal bins that are not deducible. 
\end{enumerate}

We enabled LLMs to complete the games under these different action space configurations.
To be specific, we applied supervised fine-tuning (SFT) on several well-known open-source models, including 
Meta-Llama-3-8B-Instruct~\citep{llama3modelcard}, Llama3.1-8B-Instruct~\citep{grattafiori2024llama} and Qwen2.5-7B-Instruct~\citep{qwen2.5} models. 
To prepare data for fine-tuning, we designed a planner to generate solutions under different configurations of action space. Please refer to the appendix for the details of planner design.
We varied the number of objects per game and collected trajectories for games with 4, 5, and 6 objects, totaling 250, 500, and 500 games, respectively. 
For each game, we generated five distinct solution trajectories (both optimal and near-optimal solutions), creating a large pool from which we sampled 1,000 trajectories for fine-tuning. 
Each trajectory averages around 10 steps, resulting in approximately 10,000 training samples in the chat form. 
We evaluated the models on 300 unseen games (100 games for each object count) and reported average performance across them. We also evaluated models with and without chain-of-thought (CoT) reasoning~\citep{wei2022chain}. 
The reasoning traces for the training data are generated using GPT-4o. Examples of the reasoning traces can be found in the game play section of Figure~\ref{fig:acl:game_intro}. A full example of a game play can be found in Appendix~\ref{appendix:acl:game_example}.

\section{Environment-based Verifier}
\label{sec:acl:verifier}


Inspired by the trial-and-error paradigm, we draw an analogy to how agents refine decisions through iterative interaction with their environment. Rather than training a dedicated verifier, we directly leverage environment-provided feedback as a source of verification during inference.
We evaluate whether this lightweight, training-free approach is sufficient to support LLM agents in collaborative reasoning tasks under information asymmetry.

We design a \textbf{reasoning verifier} based on environment feedback. Firstly, it is capable to examine the generated action with the game rules (e.g. physical affordance) and previous communication (e.g. redundant information sharing). Secondly, as a strategy-driven game, we implement a graph expansion algorithm to enhance agents' reasoning ability. It treats objects and bins as nodes and constraints as edges, and infers new constraints by combining existing ones. It is expanded iteratively using the transitivity of adjacent edges until no new constraints emerge. For example, objects \texttt{A} and \texttt{C} can be inferred to be on the same row if both \texttt{A} and \texttt{B}, and \texttt{B} and \texttt{C}, are known to be on the same row.
This enables the verifier to assess whether a proposed action aligns well with the current knowledge, and avoids unnecessary trials.

Further, we tested several alternatives of environment-based verifiers, and discussed their effectiveness and potential generality in Appendix~\ref{appendix:acl:verifier}.

\section{Experiments and Results}
\label{sec:acl:exp_and_results}
We conduct extensive experiments to evaluate the performance of language models in reasoning and communication under information asymmetry. Our experiments address four key research questions:

\begin{itemize} 
    \item \textbf{RQ1: Collaboration under Varying Communicative Action Space.} How do agents collaborate under information asymmetry when equipped with different communicative actions? 
    \item \textbf{RQ2: Effectiveness of Environment-Based Verification.} To what extent can environment-based verification enhance the performance? 
    \item \textbf{RQ3: Collaboration under Mismatched Communicative Action Space.} What happens when agents with mismatched communication capabilities are paired together?
    \item \textbf{RQ4: Human Preferences Toward Different Communication Behaviors.} While agents may perform well in self-play, do these behaviors align with human preferences in collaboration?
\end{itemize}

\label{sec:experiment_setup}
\subsection{Experiment Setup}

To systematically investigate our research questions, we design a series of experiments to investigate specific aspects of the collaboration:

\begin{itemize} 
    \item \textbf{Exp1: Collaboration with Different Communicative Action Spaces (RQ1)} We evaluated both closed-source language models via API calls and open-source models with supervised fine-tuning (SFT). We deploy the same model (with the same action space) to perform self-play in the game with 4, 5, and 6 objects. This setting allows direct comparison of collaboration with different communicative action space.

    \item \textbf{Exp2: Environment-Based Verifier (RQ2)} We augment the base models with an environment-based verifier, which provides binary feedback indicating whether each sampled action is valid. At each decision step, we sample 4 candidate responses using \texttt{temperature=0.2} and \texttt{top-p=0.9}, and apply the verifier to filter out invalid actions, selecting the first valid one as the final output. This setup enables a direct comparison of model performance with and without the verifier. 
    
    \item \textbf{Exp3: Collaboration with Mismatched Action Spaces (RQ3)} We select the well-performing models from those action spaces (Llama3.1-8B without CoT for \textit{no information exchange}, and Llama3.1-8B with CoT for the rest), equipped with reasoning verifier, and let them play with each other. This setting further reveals the behaviors of collaboration between agents with different communication capabilities.
    
    \item \textbf{Exp4: Human Performance (RQ4)} Once we identified the most effective configuration of communicative action space for collaborative reasoning among LLM agents, we further examined whether such models are also preferred by human users. 
    We recruited 32 participants to interact with four variants of the best-performing model. All variants used the reasoning verifier but differed in their communicative action spaces: \textit{None}, \textit{Provide Only}, \textit{Seek Only}, and \textit{Provide \& Seek}. We adopted a within-participant design crossing three game configurations (4, 6, and 8 objects) with the four model variants. Each participant completed 12 games, one under each condition, without being informed of the active variant. Each specific game-instance and model-variant pairing was evaluated by four participants. After excluding two participants who did not meet the performance threshold, we got 360 sessions in total.
    To complement quantitative metrics, we also collected qualitative feedback from participants after each game using the following three questions:
    \vspace{3pt}
    \begin{enumerate} 
        \item \textit{Do you believe the information communicated to you by the bot was useful?}
        \item \textit{Do you believe the bot effectively used the information you shared?}
        \item \textit{Do you feel confused about the bot's behavior?}
    \end{enumerate}
    \vspace{3pt}
    These questions allowed us to assess the perceived helpfulness, responsiveness, and clarity of the model’s behavior from a human-centered perspective.
\end{itemize}

\noindent \textbf{Evaluation Metrics} We employ complementary metrics to assess both effectiveness and efficiency. For effectiveness, we report the Success Rate (\texttt{SR}) and subgoal success rate (\texttt{Sub.R}) at the first attempt (Pass@1). \texttt{SR} measures the percentage of games successfully completed within a limited number of steps, while \texttt{Sub.R} reflects partial progress by capturing the proportion of objects correctly placed, even when the full game is not successfully completed.
For efficiency, we track the number of steps taken to complete each game and compare it to the optimal solution calculated by our planner. We then compute the Step Ratio (\texttt{StepR}), defined as the ratio of the number of executed steps to the number of steps in the optimal solution. 
When a verifier is applied, we additionally report the correction rate (\texttt{Corr.R}), indicating the proportion of responses corrected by the verifier. We report standard error for each metric.

\subsection{Result Analysis}

\paragraph{Effects of Communicative Action Space (RQ1)} 

\begin{table*}[t!]
\centering
\footnotesize
\setlength{\tabcolsep}{3.5pt}


\resizebox{\textwidth}{!}{%
\begin{tabular}{lrrrrrrr}
 & \multicolumn{3}{c}{\textbf{No Verifier}} & \multicolumn{4}{c}{\textbf{With Reasoning Verifier}} \\[1pt]
\textbf{Model}& SR↑(\%) & Sub.R↑(\%) & StepR↓ & \multicolumn{1}{c}{SR↑(\%)} & Sub.R↑(\%) & StepR↓ & Corr.R(\%) \\
\hline
\rowcolor{gray!20}
\multicolumn{8}{l}{\textbf{\textit{Information Providing \& Seeking}}} \\
GPT4o CoT & 51.00\spm{2.89} & 76.73\spm{1.64} & 1.92x\ssr{0.04} & 80.00\,\textcolor{red}{\scriptsize \textbf{(+29.00)}}\spm{2.31} & 91.34\spm{1.17} & 1.65x\ssr{0.03} & 17.42\spm{0.57} \\[3pt]

Llama3-8B & 13.67\spm{1.98} & 60.04\spm{1.37} & \textbf{1.47x\ssr{0.08}} & 32.33\,\textcolor{red}{\scriptsize \textbf{(+18.66)}}\spm{2.70} & 73.28\spm{1.36} & 1.40x\ssr{0.05} & 6.36\spm{0.43} \\
Llama3.1-8B & 27.33\spm{2.57} & 68.74\spm{1.44} & 1.50x\ssr{0.05} & 53.00\,\textcolor{red}{\scriptsize \textbf{(+25.67)}}\spm{2.88} & 81.05\spm{1.40} & \textbf{1.36x\ssr{0.03}} & 7.19\spm{0.44} \\
Qwen2.5-7B & 27.00\spm{2.56} & 72.31\spm{1.26} & 1.56x\ssr{0.06} & 47.00\,\textcolor{red}{\scriptsize \textbf{(+20.00)}}\spm{2.88} & 83.96\spm{1.05} & 1.45x\ssr{0.04} & 8.34\spm{0.42} \\
Llama3-8B CoT & 29.67\spm{2.64} & 62.51\spm{1.75} & 1.94x\ssr{0.09} & 70.00\,\textcolor{red}{\scriptsize \textbf{(+40.33)}}\spm{2.65} & 87.26\spm{1.34} & 1.61x\ssr{0.04} & 14.28\spm{0.58} \\
Llama3.1-8B CoT & \textbf{58.67\spm{2.84}} & \textbf{79.39\spm{1.64}} & 1.87x\ssr{0.05} & \textbf{89.33}\,\textcolor{red}{\scriptsize \textbf{(+30.66)}}\spm{1.78} & \textbf{95.64\spm{0.81}} & 1.52x\ssr{0.03} & 14.38\spm{0.59} \\
Qwen2.5-7B CoT & 44.67\spm{2.87} & 74.31\spm{1.66} & 1.91x\ssr{0.07} & 81.00\,\textcolor{red}{\scriptsize \textbf{(+36.33)}}\spm{2.26} & 92.53\spm{1.03} & 1.61x\ssr{0.03} & 14.59\spm{0.54} \\
\hline
\rowcolor{gray!20}
\multicolumn{8}{l}{\textbf{\textit{Information Providing Only}}} \\
Llama3-8B & 9.33\spm{1.68} & 62.46\spm{1.19} & 1.26x\ssr{0.08} & 27.00\,\textcolor{red}{\scriptsize \textbf{(+17.67)}}\spm{2.56} & 74.81\spm{1.23} & 1.45x\ssr{0.08} & 6.43\spm{0.38} \\
Llama3.1-8B & 26.00\spm{2.53} & \textbf{69.71\spm{1.40}} & 1.56x\ssr{0.06} & 40.33\,\textcolor{red}{\scriptsize \textbf{(+14.33)}}\spm{2.83} & 80.43\spm{1.19} & 1.43x\ssr{0.05} & 6.38\spm{0.39} \\
Qwen2.5-7B & 8.00\spm{1.57} & 51.96\spm{1.30} & \textbf{1.12x\ssr{0.02}} & 24.67\,\textcolor{red}{\scriptsize \textbf{(+16.67)}}\spm{2.49} & 67.53\spm{1.43} & \textbf{1.17x\ssr{0.04}} & 6.00\spm{0.40} \\
Llama3-8B CoT & 26.00\spm{2.53} & 60.06\spm{1.76} & 1.62x\ssr{0.10} & 56.00\,\textcolor{red}{\scriptsize \textbf{(+30.00)}}\spm{2.87} & 80.27\spm{1.54} & 1.56x\ssr{0.05} & 10.70\spm{0.56} \\
Llama3.1-8B CoT & \textbf{37.00\spm{2.79}} & 68.34\spm{1.72} & 1.70x\ssr{0.07} & \textbf{65.33}\,\textcolor{red}{\scriptsize \textbf{(+28.33)}}\spm{2.75} & \textbf{84.25\spm{1.44}} & 1.52x\ssr{0.04} & 11.54\spm{0.56} \\
Qwen2.5-7B CoT & 17.00\spm{2.17} & 52.61\spm{1.69} & 1.81x\ssr{0.10} & 38.00\,\textcolor{red}{\scriptsize \textbf{(+21.00)}}\spm{2.80} & 70.40\spm{1.69} & 1.59x\ssr{0.06} & 11.01\spm{0.53} \\
\hline
\rowcolor{gray!20}
\multicolumn{8}{l}{\textbf{\textit{Information Seeking Only}}} \\
Llama3-8B & 17.67\spm{2.20} & 63.70\spm{1.39} & 1.21x\ssr{0.05} & 37.33\,\textcolor{red}{\scriptsize \textbf{(+19.66)}}\spm{2.79} & 76.87\spm{1.30} & 1.28x\ssr{0.04} & 7.61\spm{0.47} \\
Llama3.1-8B & 33.67\spm{2.73} & 74.76\spm{1.39} & 1.50x\ssr{0.05} & 52.00\,\textcolor{red}{\scriptsize \textbf{(+18.33)}}\spm{2.88} & 84.07\spm{1.19} & 1.37x\ssr{0.03} & 6.41\spm{0.36} \\
Qwen2.5-7B & 14.00\spm{2.00} & 60.06\spm{1.29} & \textbf{1.06x\ssr{0.03}} & 24.33\,\textcolor{red}{\scriptsize \textbf{(+10.33)}}\spm{2.48} & 70.18\spm{1.29} & \textbf{1.14x\ssr{0.04}} & 3.68\spm{0.28} \\
Llama3-8B CoT & 52.33\spm{2.88} & 77.91\spm{1.64} & 1.35x\ssr{0.03} & \textbf{82.33}\,\textcolor{red}{\scriptsize \textbf{(+30.00)}}\spm{2.20} & 91.84\spm{1.15} & 1.28x\ssr{0.03} & 10.15\spm{0.47} \\
Llama3.1-8B CoT & \textbf{56.67\spm{2.86}} & \textbf{81.14\spm{1.45}} & 1.42x\ssr{0.04} & 82.67\,\textcolor{red}{\scriptsize \textbf{(+26.00)}}\spm{2.19} & \textbf{93.42\spm{0.93}} & 1.25x\ssr{0.02} & 10.59\spm{0.47} \\
Qwen2.5-7B CoT & 39.33\spm{2.82} & 69.57\spm{1.77} & 1.59x\ssr{0.05} & 76.67\,\textcolor{red}{\scriptsize \textbf{(+37.34)}}\spm{2.44} & 89.90\spm{1.27} & 1.35x\ssr{0.03} & 14.44\spm{0.53} \\
\hline
\rowcolor{gray!20}
\multicolumn{8}{l}{\textbf{\textit{No Information Exchange}}} \\
Llama3-8B & 81.33\spm{2.25} & 92.79\spm{0.98} & \textbf{1.88x\ssr{0.04}} & \textbf{97.67}\,\textcolor{red}{\scriptsize \textbf{(+16.34)}}\spm{0.87} & \textbf{99.39\spm{0.25}} & \textbf{1.61x\ssr{0.03}} & 11.66\spm{0.57} \\
Llama3.1-8B & \textbf{84.67\spm{2.08}} & \textbf{96.13\spm{0.62}} & 2.08x\ssr{0.04} & 94.33\,\,\,\textcolor{red}{\scriptsize \textbf{(+9.66)}}\spm{1.33} & 98.50\spm{0.42} & 1.68x\ssr{0.03} & 11.09\spm{0.60} \\
Qwen2.5-7B & 34.67\spm{2.75} & 76.90\spm{1.24} & 1.91x\ssr{0.07} & 61.33\,\textcolor{red}{\scriptsize \textbf{(+26.66)}}\spm{2.81} & 88.21\spm{1.03} & 1.79x\ssr{0.05} & 10.28\spm{0.49} \\
Llama3-8B CoT & 33.33\spm{2.72} & 66.78\spm{1.79} & 2.21x\ssr{0.10} & 73.00\,\textcolor{red}{\scriptsize \textbf{(+39.67)}}\spm{2.56} & 90.51\spm{1.06} & 1.91x\ssr{0.06} & 17.14\spm{0.62} \\
Llama3.1-8B CoT & 38.33\spm{2.81} & 73.83\spm{1.53} & 2.16x\ssr{0.08} & 54.33\,\textcolor{red}{\scriptsize \textbf{(+16.00)}}\spm{2.88} & 83.37\spm{1.28} & 1.82x\ssr{0.06} & 14.35\spm{0.58} \\
Qwen2.5-7B CoT & 39.67\spm{2.82} & 71.02\spm{1.73} & 2.43x\ssr{0.10} & 65.33\,\textcolor{red}{\scriptsize \textbf{(+25.66)}}\spm{2.75} & 85.99\spm{1.31} & 1.96x\ssr{0.06} & 15.79\spm{0.58} \\
\hline
\end{tabular}
}
\caption{Performance comparison of models with and without verifier assistance across four communicative action space configurations. In each game, both agents are assigned the same action space.}
\label{tab:acl:main_result}
\vspace{-5pt}
\end{table*}
 
We observe that the effect of communicative action space depends on the model and reasoning configuration. In the strongest Llama 3.1-8B CoT setting, \emph{Provide \& Seek} achieves the highest Success Rate (\texttt{SR}). Across matched settings, \emph{Information Seeking Only} also generally outperforms \emph{Information Providing Only}, suggesting that targeted queries are more effective than unprompted sharing. Queries allow agents to request missing information directly, whereas proactive sharing often introduces constraints that are redundant or irrelevant to the partner. The corresponding Step Ratio (\texttt{StepR}) rankings are less consistent, indicating that higher task success does not always translate into greater step efficiency.

Surprisingly, disabling communication altogether (\emph{`No Information Exchange’’}) ranks high for some variants. Without communication and reasoning process through CoT, the task reduces to pure object manipulation and random guessing; 
Llama variants exploit this by memorizing high-probability transition patterns, achieving a success rate greater than 81\%. Qwen, on the other hand, does not show the same effect, suggesting the phenomenon is model–specific rather than an intrinsic property of the environment.

\paragraph{Effects of Verifier Assistance (RQ2)}
Holding the communicative action space and reasoning setting fixed, adding the reasoning verifier produces absolute \texttt{SR} gains of 10 to 40 percentage points, as shown in red in Table~\ref{tab:acl:main_result}. The improvement is observed across action spaces, indicating that verifier-assisted action selection helps agents act more consistently with the current game state and constraints.


\paragraph{Error Analysis on RQ1 and RQ2}
\newcommand{\smallpct}{{\small\%}}
\begin{table*}[t]
\vspace{-10pt}
\centering
\setlength{\tabcolsep}{1pt}
\resizebox{0.99\linewidth}{!}{
\begin{tabular}{lcccc|cccc}
\multirow{2}{*}{Category} & \multicolumn{4}{c}{\textbf{No Verifier}} & \multicolumn{4}{c}{\textbf{With Reasoning Verifier}} \\
 & \textbf{Provide\,\&\,Seek} & \textbf{Provide Only} & \textbf{Seek Only} & \textbf{None} & \textbf{Provide\,\&\,Seek} & \textbf{Provide Only} & \textbf{Seek Only} & \textbf{None} \\
\hline
Format Following & 0.19\smallpct & 0.59\smallpct & 0.00\smallpct & 0.00\smallpct & 0.08\smallpct & 0.48\smallpct & 0.02\smallpct & 0.00\smallpct \\
\hline
\rowcolor{gray!20}
\textbf{Physical Understanding} & & & & & & & & \\
Object not in source bin & 8.34\smallpct/11.27\smallpct & 4.17\smallpct/7.75\smallpct & 10.81\smallpct/15.16\smallpct & 1.47\smallpct/1.60\smallpct & 2.92\smallpct/4.06\smallpct & 1.94\smallpct/3.39\smallpct & 4.89\smallpct/7.05\smallpct & 1.11\smallpct/1.19\smallpct \\
Source bin not reachable & 1.43\smallpct/1.94\smallpct & 0.77\smallpct/1.43\smallpct & 0.84\smallpct/1.18\smallpct & 0.02\smallpct/0.02\smallpct & 0.97\smallpct/1.34\smallpct & 1.00\smallpct/1.75\smallpct & 0.24\smallpct/0.35\smallpct & 0.02\smallpct/0.03\smallpct \\
Dest. bin not reachable & 4.27\smallpct/5.77\smallpct & 3.83\smallpct/7.11\smallpct & 3.67\smallpct/5.15\smallpct & 0.00\smallpct/0.00\smallpct & 2.36\smallpct/3.28\smallpct & 2.50\smallpct/4.37\smallpct & 2.06\smallpct/2.98\smallpct & 0.00\smallpct/0.00\smallpct \\
Source \& destination same & 0.19\smallpct/0.26\smallpct & 0.70\smallpct/1.30\smallpct & 0.81\smallpct/1.13\smallpct & 0.02\smallpct/0.02\smallpct & 0.18\smallpct/0.26\smallpct & 0.60\smallpct/1.04\smallpct & 0.24\smallpct/0.35\smallpct & 0.02\smallpct/0.03\smallpct \\
\hline
\rowcolor{gray!20}
\textbf{Communication} & & & & & & & & \\
Redundant knowl. sharing & 14.04\smallpct/59.86\smallpct & 33.20\smallpct/79.35\smallpct & 0.08\smallpct/0.89\smallpct & -- & 9.60\smallpct/39.44\smallpct & 27.61\smallpct/69.37\smallpct & 0.10\smallpct/0.83\smallpct & -- \\
No share after seek & 0.30\smallpct/12.99\smallpct & -- & 0.93\smallpct/9.65\smallpct & -- & 0.14\smallpct/4.00\smallpct & -- & 0.16\smallpct/1.31\smallpct & -- \\
Wrong share after seek & 0.04\smallpct/0.19\smallpct & -- & 0.00\smallpct/0.00\smallpct & -- & 0.04\smallpct/0.17\smallpct & -- & 0.00\smallpct/0.00\smallpct & -- \\
Seek known object & 0.34\smallpct/14.94\smallpct & -- & 1.21\smallpct/12.54\smallpct & -- & 0.02\smallpct/0.57\smallpct & -- & 0.24\smallpct/1.96\smallpct & -- \\
\hline
\rowcolor{gray!20}
\textbf{Task Reasoning} & & & & & & & & \\
Wrong rule understanding & 28.22\smallpct/38.15\smallpct & 19.07\smallpct/35.46\smallpct & 27.40\smallpct/38.41\smallpct & 27.35\smallpct/29.79\smallpct & 14.12\smallpct/19.62\smallpct & 13.13\smallpct/22.94\smallpct & 17.53\smallpct/25.27\smallpct & 19.31\smallpct/20.67\smallpct \\
Wrong random guessing & 8.64\smallpct/11.64\smallpct & 4.25\smallpct/7.90\smallpct & 5.76\smallpct/8.07\smallpct & 16.11\smallpct/17.54\smallpct & 6.12\smallpct/8.51\smallpct & 4.90\smallpct/8.56\smallpct & 3.57\smallpct/5.14\smallpct & 14.77\smallpct/15.81\smallpct \\
\textbf{No Error} & 44.29\smallpct & 38.80\smallpct & 59.61\smallpct & 55.92\smallpct & 67.19\smallpct & 51.37\smallpct & 75.68\smallpct & 65.60\smallpct \\
\textbf{Total Actions} & 6704 & 7293 & 6427 & 5308 & 4867 & 5881 & 4991 & 4143 \\
\hline
\end{tabular}}
\caption{Error analysis across different communicative action spaces with/without verifier.}
\label{tab:acl:error_analysis}
\vspace{-5pt}
\end{table*}

To examine the behaviors underlying these aggregate results, we analyze errors for the best-performing LLaMA3.1-8B configuration under each action space. We use the variant without CoT for \emph{None} and the variant with CoT for the other three action spaces. The same configurations are used in subsequent experiments. Appendix~\ref{appendix:acl:error_taxonomy} provides the complete error taxonomy. Each cell in Table~\ref{tab:acl:error_analysis} reports two rates:
\begin{itemize}
\item \textsc{(Errors/Total Steps)\%}: the percentage of all steps assigned a given error label.
\item \textsc{(Errors/Relevant Action Type)\%}: the percentage of actions of the relevant type, such as move, provide, or seek, assigned that label.
\end{itemize}

For example, under \emph{Provide \& Seek} without the verifier, the \emph{Wrong Rule Understanding} rates are 28.22\% and 38.15\%. These values indicate that the error occurs in 28.22\% of all 6{,}704 steps and 38.15\% of the 4{,}959 move actions.\footnote{The total number of move actions is not shown in the table.} Because the error labels are not mutually exclusive, a single action may contribute to multiple categories.

Without the verifier, agents exhibit broadly similar rates of rule-understanding errors across action spaces despite substantial differences in task success. The relatively high success rate under \emph{None} therefore appears to result from repeated manipulation attempts and the absence of communication-related errors, rather than stronger task reasoning. Meanwhile, communication is also frequently inefficient. Among information-providing actions, 59.86\% under \emph{Provide \& Seek} and 79.35\% under \emph{Provide Only} are labeled as redundant. Agents also have difficulty responding to requests and seeking necessary information. Providing information only in response to a request reduces redundancy, although its overall performance remains below that of the full bidirectional action space.

Adding the reasoning verifier consistently reduces error rates and the total number of actions. The reduction is particularly pronounced for \emph{Wrong Rule Understanding}, which directly reflects the verifier's intended role. The verifier also reduces communication errors, leading to more efficient and relevant information exchange.

Together, these results distinguish the roles of communicative action space and verification. The action space determines what information agents can exchange, while the verifier improves how reliably agents use those actions and follow the game constraints. Full bidirectional communication provides the strongest overall action space, and verifier assistance further improves both task performance and rule compliance.

We further evaluate whether these findings extend beyond the game sizes used during fine-tuning. Models fine-tuned only on games with 4--6 objects are evaluated on unseen games with 7 and 8 objects (Appendix~\ref{appendix:acl:generalization}, Table~\ref{tab:acl:generalization}). Although absolute performance decreases as the number of objects increases, the ordering across communicative action spaces remains largely unchanged. The verifier also retains substantial gains, increasing \texttt{SR} from 31.00\% to 76.00\% under \emph{Provide \& Seek} on the held-out sizes. These results show some extent of generalizability that the environment-based verifier can bring.

\ifshowold
\begin{figure*}[t!]
  \centering
  \begin{minipage}[b]{0.51\textwidth}
    \centering
    \small
    \setlength{\tabcolsep}{1.9pt}
    \begin{tabular}{cccc}
    \textbf{Act. Mode 1} & \textbf{Act. Mode 2} & \textbf{SR\textuparrow(\%)} & \textbf{StepR\textdownarrow} \\
    \hline
    Provide\,\&\,Seek  & Provide Only & 67.33\spm{2.71} & 1.57x\ssr{0.04} \\
    Provide\,\&\,Seek  & Seek Only    & 78.67\spm{2.37} & 1.33x\ssr{0.02} \\
    Provide\,\&\,Seek  & None         & 78.67\spm{2.37} & 1.68x\ssr{0.04} \\
    Provide Only       & Seek Only    & 41.00\spm{2.84} & 1.63x\ssr{0.07} \\
    \hline
    \end{tabular}
    \captionof{table}{Comparison across different action space combinations. All variants are equipped with reasoning verifier.}
    \label{tab:acl:mismatch_action_space}
  
    \vspace{20pt}
  
    \centering
    \small
    \setlength{\tabcolsep}{2pt}
    \begin{tabular*}{\linewidth}{@{\extracolsep{\fill}}lccc}
        \textbf{Action Mode}    & \textbf{SR\textuparrow(\%)} & \textbf{Sub.R\textuparrow(\%)} & \textbf{Steps\textdownarrow}\\
        \hline
        {Provide\,\&\,Seek}  & 92.22\spm{2.62} & 97.41\spm{1.02} & 18.99\spm{0.58} \\
        {Provide Only}       & 77.78\spm{4.02} & 93.19\spm{1.60} & 22.07\spm{0.81} \\
        {Seek Only}          & 87.78\spm{2.98} & 95.19\spm{1.39} & 19.10\spm{0.57} \\
        {None}               & 91.11\spm{2.74} & 98.06\spm{0.70} & 18.76\spm{0.58} \\
        \hline
    \end{tabular*}
    \captionof{table}{Performance of collaboration between best agents and 30 human participants.}
    \label{tab:acl:human_result}
  \end{minipage}%
  \hfill
  \begin{minipage}[b]{0.47\textwidth}
    \centering
    \includegraphics[width=\linewidth]{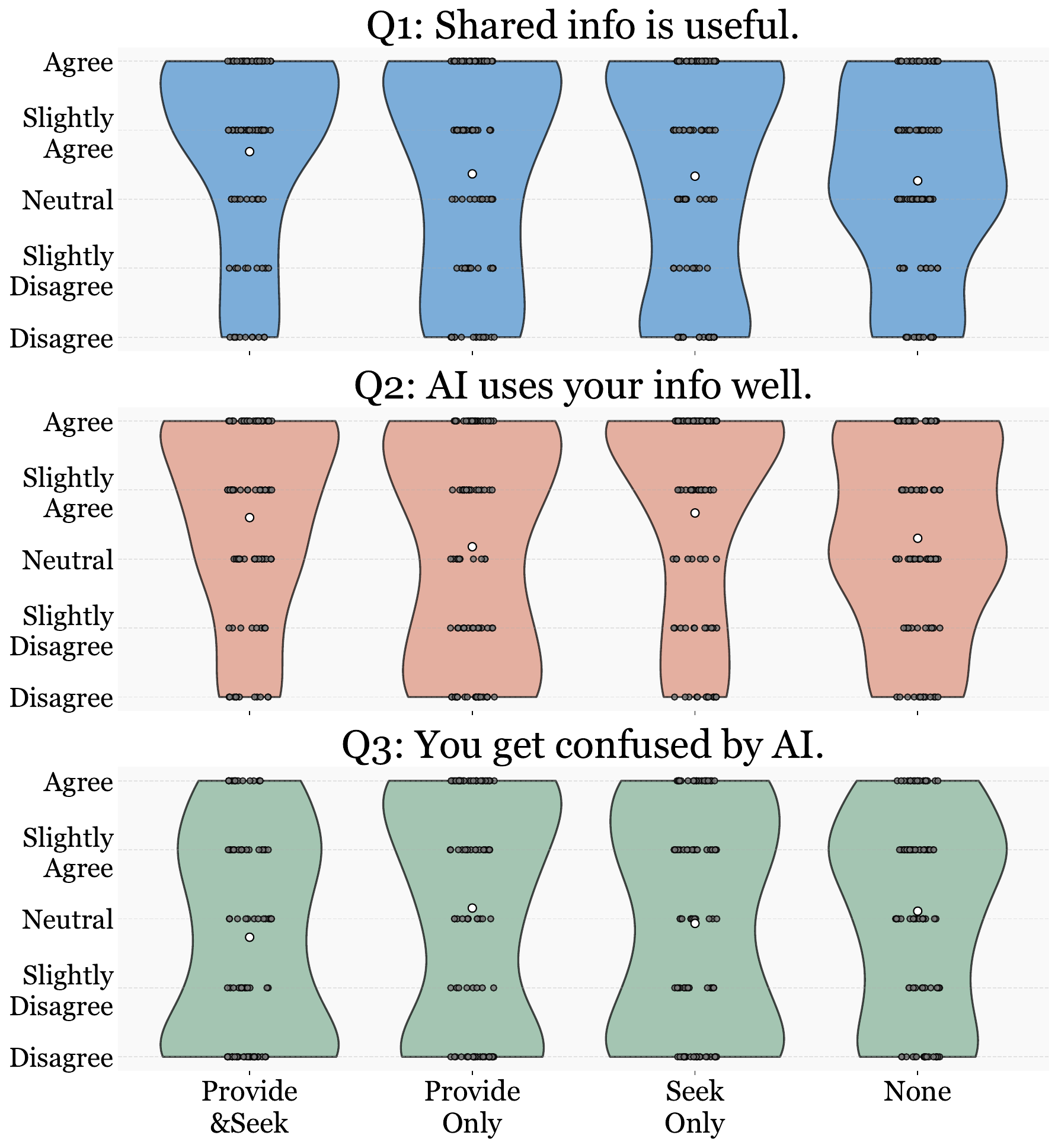}
    \vspace{-15pt}
    \captionof{figure}{Distributions of answers with 360 data points from 30 human participants.}
    \label{fig:acl:human_study_result}
  \end{minipage}
  \end{figure*}
\else
\begin{table}[t!]
  \centering
  \small
  \resizebox{\linewidth}{!}{%
  \begin{tabular}{cccc}
  \textbf{Act. Mode 1} & \textbf{Act. Mode 2} & \textbf{SR\textuparrow(\%)} & \textbf{StepR\textdownarrow} \\
  \hline
  Provide\,\&\,Seek  & Provide Only & 67.33\spm{2.71} & 1.57x\ssr{0.04} \\
  Provide\,\&\,Seek  & Seek Only    & 78.67\spm{2.37} & 1.33x\ssr{0.02} \\
  Provide\,\&\,Seek  & None         & 78.67\spm{2.37} & 1.68x\ssr{0.04} \\
  Provide Only       & Seek Only    & 41.00\spm{2.84} & 1.63x\ssr{0.07} \\
  \hline
  \end{tabular}}
  \caption{Comparison across different action space combinations. All variants are equipped with reasoning verifier.}
  \label{tab:acl:mismatch_action_space}
\end{table}
\fi
  
\paragraph{Mismatched Action Spaces (RQ3)} 
We further examine model behavior in scenarios where two agents are assigned different communicative action spaces. Specifically, we evaluate four asymmetric pairings of action spaces, as shown in Table~\ref{tab:acl:mismatch_action_space}. We exclude the pair Provide only vs. None, as it is functionally similar to \textit{Provide \& Seek} vs. \textit{None} in information transfer. In both cases, one agent is unable to respond to queries. Similarly, \textit{Seek only} vs. \textit{None} is the same as \textit{None} vs. \textit{None}, since the \textit{None} agent cannot initiate or respond to communication. 

Across the evaluated pairings, we observe a consistent drop in performance when agents have mismatched communicative abilities. This finding highlights the importance of aligning communication protocols in collaborative tasks and provides insights into designing agent communication strategies in multi-agent systems. Notably, in the last two pairings listed in Table~\ref{tab:acl:mismatch_action_space}, agents are sometimes forced to rely on random guessing. For instance, a \textit{Seek only} agent cannot proactively share unless asked, yet its \textit{Provide only} partner lacks the ability to initiate queries. These structural mismatches lead to particularly sharp declines in task success, strengthening the need for matched-communication agent design.

\paragraph{Human Preference (RQ4)}
We follow the model selection in Table~\ref{tab:acl:mismatch_action_space} from \textbf{RQ3} and recruit human participants to play games with them. 

As shown in Table~\ref{tab:acl:human_result}, Provide\,\&\,Seek achieves the highest success rate (92.2\%) in collaboration with human participants, followed closely by None (91.1\%), while Provide Only trails substantially (77.8\%). Although Provide\,\&\,Seek and None do not have a significant difference in steps per game (18.99 vs.\ 18.76, $p=.73$), they spend these steps very differently: the None agent resorts to trial-and-error, nearly doubling its rejected placement attempts (3.64 vs.\ 1.92 per game, $p<.001$), whereas the Provide\,\&\,Seek agent devotes a comparable share of its turns to communication (1.9 communicative acts per game), replacing failed physical attempts with information exchange almost one for one. This suggests that agents trained with communicative actions produce fewer erroneous physical actions and better acquire the underlying task strategy, and leverage effective communication to avoid erroneous physical actions.

In addition to quantitative metrics, we also conduct qualitative analyses to better understand the perceived helpfulness, responsiveness, and clarity of the models from a human-centered perspective. Figure~\ref{fig:acl:human_study_result} presents the distributions of participant responses across these three dimensions. Participants rate the information communicated by Provide\,\&\,Seek agents as significantly more useful than in the None condition ($p=.020$), and Seek Only agents are rated significantly higher than None on making effective use of human-shared information ($p=.040$). One possible interpretation is that, under the same training budget, the Seek Only agent learns a more focused information-seeking policy: it asks selectively for information it needs, making its use of human-provided information more salient to participants. Interestingly, in the None condition for Q1, participants often interpret the agents' physical actions as implicit communication from which they base their evaluations. Notably, participants attempted to communicate at the same rate in the None condition as in Provide\,\&\,Seek (3.0 acts per game, $p=.94$), despite never receiving a response, which suggests a persistent human demand for communication that non-communicative agents leave unmet.

Consistently, we observe \textbf{significant differences in the perceived clarity} of agent behavior: participants report the most confusion in the None condition and the least with Provide\,\&\,Seek ($p=.045$). This indicates that communication improves perceived clarity without increasing the number of steps. Finally, Provide Only performs worse than all other variants, driven largely by the hardest games (46.7\% success with 8 objects). A possible indication is that useful providing is harder to learn than useful seeking: to provide information effectively, the agent may infer what the partner already knows, identify what is still useful, and decide when to volunteer it, whereas seeking directly exposes the agent's own information needs. This may also explain why Provide\,\&\,Seek performs best, since seeking can expose missing information and help guide what should be provided. These results highlight that beyond task success, \textbf{proactive and transparent communication plays a critical role in improving perceived clarity in human-agent interaction}.

\label{sec:conclusion}
\section{Conclusion}


We adapted Einstein's Puzzle into a tabletop environment for studying collaboration under information asymmetry, where two agents differ in both knowledge and reachable workspace. Agents that can both seek and provide information collaborate most effectively for most cases, and mismatched communicative abilities degrade performance sharply. Our central finding, however, is a dissociation rather than a ranking. Agents without communication can achieve high task success through repeated physical attempts, but their trajectories contain more rule-inconsistent and rejected actions. Human participants also report greater confusion in this condition despite similar step counts. These results show that task success alone does not capture behavioral reliability or interaction quality. Our environment-based verifier addresses part of the gap: useful verification signals can be read directly off the environment without reward engineering or extra training, improving constraint-consistent action selection and task completion.

\clearpage

\section*{Acknowledgments}
This work has benefited from the Microsoft Accelerate Foundation Models Research (AFMR) grant program. We
thank all the human-study participants for their contributions.

\bibliography{custom}
\bibliographystyle{colm2026_conference}

\appendix

\newpage

\section*{Appendix}

We include the following contents as our supplementary materials.

\begin{itemize}[leftmargin=20pt, itemsep=3pt, parsep=0pt, topsep=0pt, partopsep=0pt] 
    \item \hyperref[appendix:acl:game_example]{\textbf{A. \textbf{Example of Game Play}}} — An illustrative walkthrough of a full game session demonstrating player turns, actions, and reasoning strategies.
    
    \item \hyperref[appendix:acl:data_generation]{\textbf{B. \textbf{Data Generation}}} — A detailed description of how we generate trajectories using a planner with perspective-taking, inference, and communication modeling.

    \item \hyperref[appendix:acl:verifier]{\textbf{C. \textbf{Ablation on Verifiers}}} - A detailed explanation and experiment on different designs of verifiers, which is potentially generalizable to other simulated environments.

    \item \hyperref[appendix:acl:generalization]{\textbf{D. \textbf{Out-of-Distribution Generalization}}} - Evaluation of models fine-tuned on 4--6 object games on unseen games with 7 and 8 objects.
    
    \item \hyperref[appendix:acl:param_details]{\textbf{E. \textbf{Experiment Details}}} — Technical and procedural specifications of model training, deployment, and human evaluation setup.
    
    \item \hyperref[appendix:acl:ethics]{\textbf{F. \textbf{Code of Ethics}}} — Statement of ethical compliance, consent protocol, and risk mitigation measures approved by the IRB.

    \item \hyperref[appendix:acl:limitation]{\textbf{G. \textbf{Limitation}}} - About generalizability of findings on communication and verification.

    \item \hyperref[appendix:acl:error_taxonomy]{\textbf{H. \textbf{Error Taxonomy}}} - A detailed explanation on the error types we analyze in \textbf{Exp2} in section~\ref{sec:acl:exp_and_results}.


    \item \hyperref[appendix:acl:interface_v2]{\textbf{I. \textbf{Human Study Interface}}} — Screenshot and description of the web interface and tutorial used for guiding participants in the human study.
    
    
    \item \hyperref[appendix:acl:prompts]{\textbf{J. \textbf{Prompts Used}}} — A comprehensive collection of system and user prompts used across training, evaluation, and reasoning trace generation for different agent configurations.

\end{itemize}


In terms of generative AI usage, we use it for purely improving the language of the paper.

\section{Example of Game Play}
\label{appendix:acl:game_example}
\begin{figure*}[t!]
    \centering
    \includegraphics[width=\linewidth]{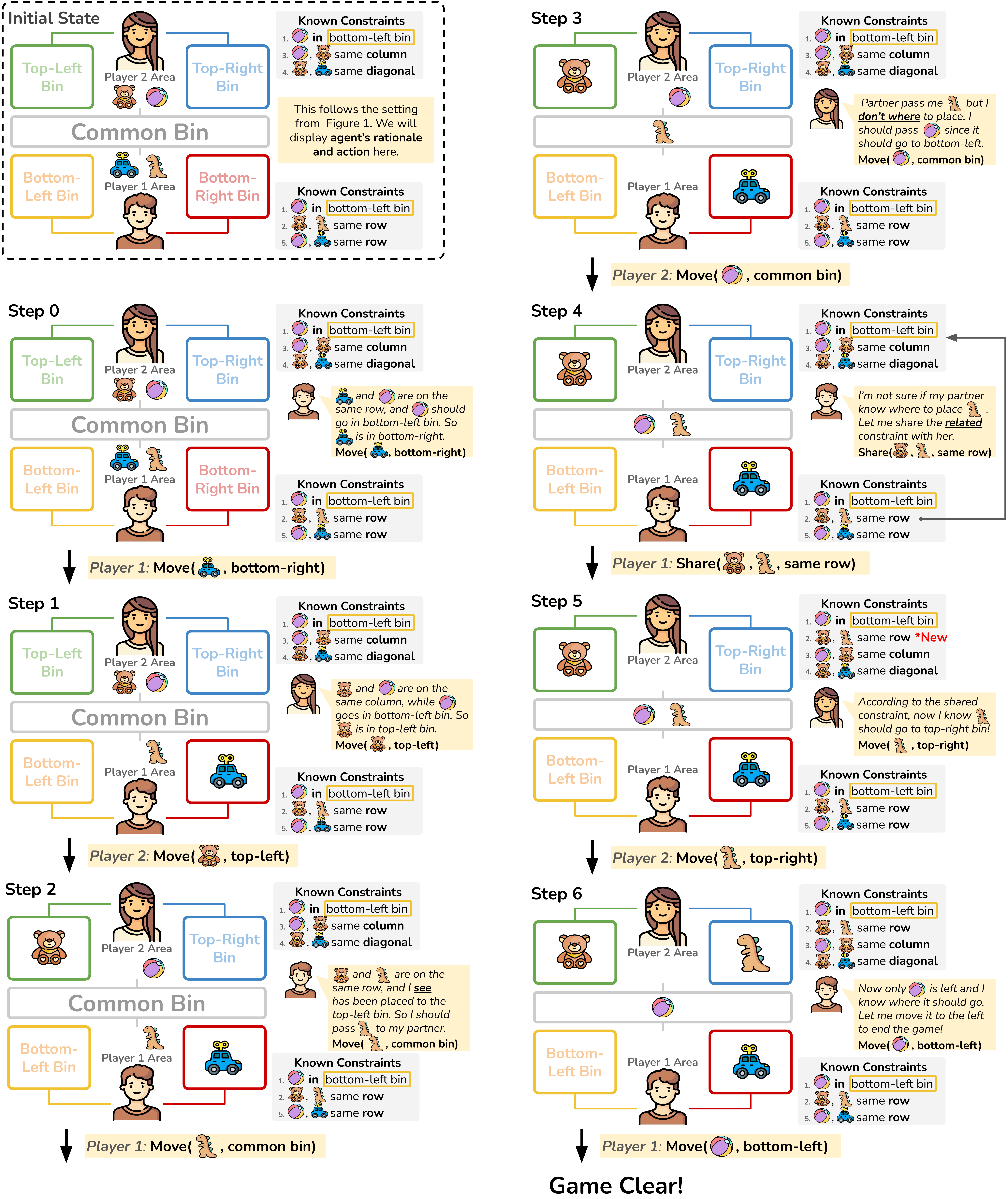}
    \caption{Full game playthrough with actions and rationales. The game begins with Player 1, and the two players take turns performing physical moves, sharing information, or asking questions until all objects are correctly placed.}
    \label{fig:acl:game_example}
\end{figure*}

To provide a clear understanding of the Einstein Puzzle gameplay, we illustrate a complete game session in Figure~\ref{fig:acl:game_example}, building on the initial setup introduced in Figure~\ref{fig:acl:game_intro}. This example demonstrates how players are expected to take turns, make decisions, and communicate throughout the game. We hope this serves as a helpful reference for understanding the nature of the task, as well as the types of reasoning and interaction involved.

It is worth noting that the \texttt{ask} action is not utilized in this example. This is primarily due to the relative simplicity of the case. The \texttt{ask} action tends to be used more frequently in scenarios involving a larger number of objects, or when agents need to inquire about specific object-related information.

\section{Data Generation}
\label{appendix:acl:data_generation}
\subsection{Perspective Taking}
To prepare data for fine-tuning, we designed a planner to generate solutions under different configurations of action space. 

The planner operates from the perspective of a single player in each turn, selecting moves based on the player's limited knowledge of the constraints and their specific communicative action space. It performs breadth-first search to explore possible action sequences until a valid solution is found. Throughout the search process, the planner maintains an up-to-date representation of the player’s knowledge, incorporating both the communication history and the current positions of objects on the board. Based on this evolving knowledge state, it selects valid next actions—such as sharing constraints that have not yet been communicated, or asking about objects that remain unknown from the player’s perspective.

To improve search efficiency, we incorporate inferred knowledge into action selection, which is also used by the reasoning verifier introduced in Section~\ref{sec:acl:verifier}. This inferred knowledge enables the planner to determine whether:
\begin{enumerate}[leftmargin=20pt, itemsep=1pt, parsep=0pt, topsep=0pt, partopsep=0pt]
\item The goal of an object is already \textbf{known}, in which case redundant information-seeking is avoided, and a valid move (placing the object into its goal bin, if reachable) is added;
\item The goal of an object is still \textbf{unknown}, in which case the agent may ask for information about that object.
\end{enumerate}
By tracking communication history, the planner can also avoid redundant sharing by identifying which constraints have already been communicated.

If no valid move is available in a given search step, we manually add a \texttt{pass} action to allow the player to skip their turn. This mechanism is important when the two players take different numbers of actions to complete a task. Any trajectory ending with two consecutive passes is pruned to avoid unnecessary stalling.

In scenarios where no communication is allowed, or where information flow is unidirectional, the agent may have to randomly guess the goals of some objects. This is triggered only after all valid actions have been exhausted, before skipping their turn. The agent will then attempt to place reachable, unplaced objects into all reachable bins (including the common bin), one by one, until the correct placement is accepted by the environment. Since the environment prevents invalid placements, this process can still lead to a valid solution. However, agents are discouraged from guessing prematurely and are designed to prioritize valid actions before resorting to this strategy.


\subsection{Optimal And Near-Optimal Trajectories}
We collect both optimal trajectories and near-optimal trajectories that are only one or two steps longer but follow alternative strategies. Including near-optimal trajectories ensures that both information-providing and information-seeking actions are adequately represented. Restricting the demonstrations to optimal trajectories would overrepresent unprompted information sharing, which often produces shorter solutions.

In scenarios involving random guessing, the optimal trajectories are those in which agents correctly guess the target location on the first attempt. However, such demonstrations provide little guidance for learning robust guessing strategies. To address this, we also include trajectories where agents try multiple possible bins, which we observe being learned by several fine-tuned models in Table~\ref{tab:acl:main_result}.
\section{Ablation on Environment-Based Verifiers}
\label{appendix:acl:verifier}
\begin{figure*}[t!]
    \centering
    \includegraphics[width=\linewidth]{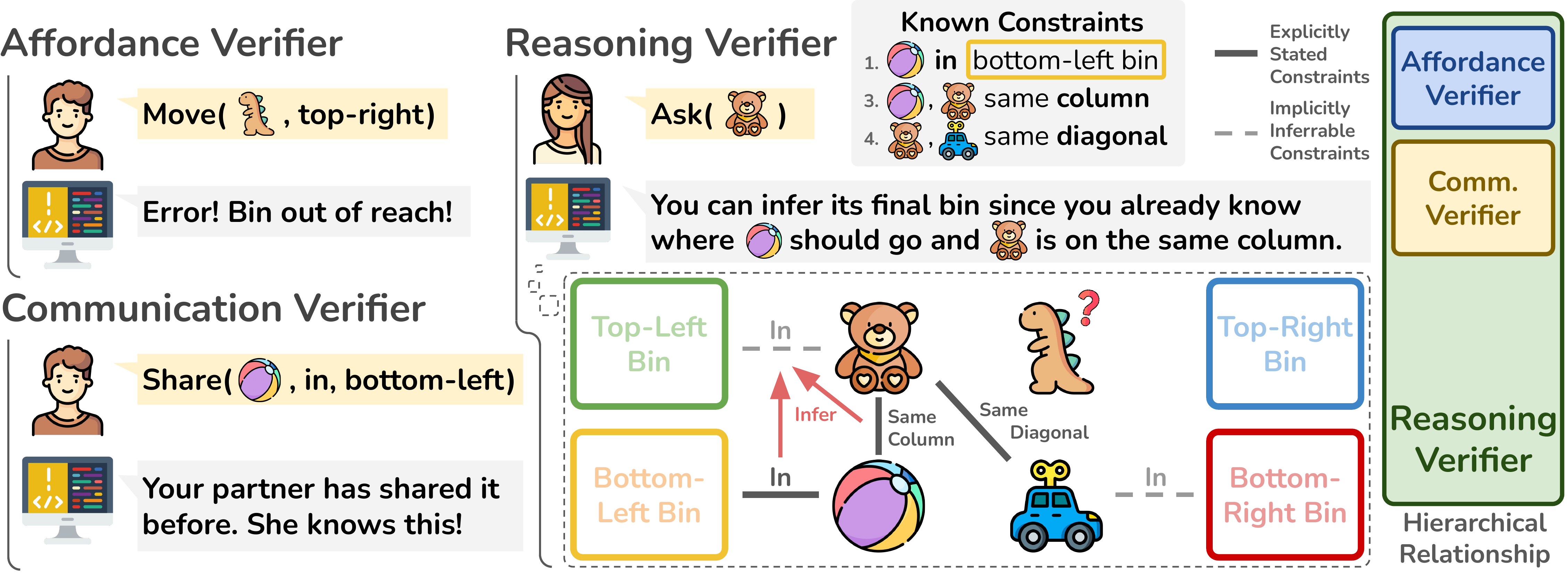}
    \caption{Illustration of three types of verifications we consider. Following the same setup as we showed in Figure 1, the game environment supports providing feedback related to action affordance, communication and strategies, which can be directly used as verifiers for agents' decisions.}
    \label{fig:acl:verifier}
\end{figure*}

\subsection{Verifier Design}
In addition to the reasoning verifier introduced in Section~\ref{sec:acl:verifier}, we design two more types of verifiers as is illustrated in Figure~\ref{fig:acl:verifier}. 

\begin{enumerate}[leftmargin=10pt, itemsep=3pt, parsep=0pt, topsep=0pt, partopsep=0pt]
    \item \textbf{Affordance Verifier}: The environment inherently enforces physical rules that govern the actions an agent can take under different conditions. In our game, this refers to whether the selected action is executable, such as whether an object or bin is reachable, or whether the chosen object is in the correct source bin. This helps validate the agent's decision from the perspective of physical feasibility. 
    \item \textbf{Communication Verifier}: In multi-agent environments, other agents can be viewed as part of the environment, and their interactions can provide additional feedback. In our game, the communication verifier assesses whether the communicative action selected by the agent is meaningful. For example, it identifies when an agent shares already-known knowledge, repeats sharing an existing constraint, or asks about an object that has already been placed.
\end{enumerate}

These three verifiers (including reasoning verifier) are hierarchically related in our design. The affordance and communication verifiers address physical actions and communication, respectively, while the reasoning verifier builds upon both and extends coverage through inferred knowledge. 

We note that the three verifiers differ substantially in how portable they are, and we distinguish them explicitly to avoid overclaiming. The affordance verifier is widely applicable and available in most environments. 
The communication verifier is similarly accessible whenever the task involves communication. 
Among the three, the reasoning verifier performs best but is also the least portable. It expands an explicit symbolic constraint graph, so adapting it to a new domain requires algorithms tailored to that domain rather than simple configuration. We therefore view environment-based verification as a general design principle: verification signals should be derived from action affordances, communication validity, and consistency with observable state. This principle may extend to other collaborative environments. 

\subsection{Ablation Study}

We follow the same setting of Exp2 (see Section~\ref{sec:acl:exp_and_results}) with two additional verifiers that separately target action affordance and communication. We compare performance across different verifier settings with Llama3.1-8B CoT model as base, and assess whether the environment-based verifier is potentially generalizable to other environments.



\begin{table*}[t!]
\centering
\small
\setlength{\tabcolsep}{3pt}
\resizebox{\textwidth}{!}{%
\begin{tabular}{lcccccccc}
\textbf{Action Mode}&
\multicolumn{2}{c}{\textbf{No Verifier (Pass@1)}} & 
\multicolumn{2}{c}{\textbf{Affordance Verifier}} & 
\multicolumn{2}{c}{\textbf{Communication Verifier}} & 
\multicolumn{2}{c}{\textbf{Reasoning Verifier}} \\
& SR↑(\%) & StepR↓ & SR↑(\%) & StepR↓ & SR↑(\%) & StepR↓ & SR↑(\%) & StepR↓ \\
\hline
Provide\,\&\,Seek & 58.67\spm{2.84} & 1.87x\ssr{0.05} & 65.67\spm{2.74} & 1.83x\ssr{0.05} & 65.00\spm{2.75} & 1.74x\ssr{0.05} & 89.33\spm{1.78} & 1.52x\ssr{0.03}   \\
Provide Only      & 37.00\spm{2.79} & 1.70x\ssr{0.07} & 38.67\spm{2.81} & 1.72x\ssr{0.06} & 46.00\spm{2.88} & 1.65x\ssr{0.05} & 65.33\spm{2.75} & 1.52x\ssr{0.04}    \\
Seek Only        & 56.67\spm{2.86} &  1.42x\ssr{0.04} & 62.33\spm{2.80} & 1.43x\ssr{0.04} & 55.33\spm{2.87} & 1.40x\ssr{0.04} & 82.67\spm{2.19} & 1.25x\ssr{0.02}   \\
None       & 38.33\spm{2.81} & 2.16x\ssr{0.08} & 42.67\spm{2.86} & 2.18x\ssr{0.09} & - & - & 54.33\spm{2.88} & 1.82x\ssr{0.06}   \\
\hline
\end{tabular}}
\caption{Performance under different verification settings tested on Llama3.1-8B model with CoT.}
\label{tab:acl:ablation_verifier}
\end{table*}


As is shown in Table \ref{tab:acl:ablation_verifier}, the affordance verifier evaluates physical preconditions of actions and improves success rates (\texttt{SR}) by 2–7\%. The communication verifier filters out redundant or uninformative exchanges, contributing up to a 9\% \texttt{SR} increase in task completion. Importantly, these verifiers operate solely on feedback from the environment and interaction history, requiring no additional training or computational overhead.

We argue that such a mechanism \textbf{offers a promising alternative} to recent agent modeling approaches, especially in simulated environments where rich, structured feedback is readily available.
These results highlight the environment's broader role as a source of lightweight, model-free verification signals that can guide agent behavior.

\section{Out-of-Distribution Generalization}
\label{appendix:acl:generalization}

To assess whether our findings depend on the game sizes observed during fine-tuning, we fine-tune all models exclusively on games with 4, 5, and 6 objects and evaluate them on games with the held-out sizes of 7 and 8 objects. Table~\ref{tab:acl:generalization} reports these results alongside the in-distribution results for reference. Although absolute performance decreases as the number of objects increases, as expected from the combinatorial growth of the constraint graph, the relative ordering of the communicative action spaces remains largely unchanged. The reasoning verifier also continues to yield substantial gains on the larger, unseen game sizes. These results suggest that the trends reported in Section~\ref{sec:acl:exp_and_results} extend to more complex games within the same task family.
\begin{table*}[t!]
\centering
\small
\setlength{\tabcolsep}{4pt}
\resizebox{\textwidth}{!}{%
\begin{tabular}{llccccc}
& & \multicolumn{2}{c}{\textbf{No Verifier}} & \multicolumn{2}{c}{\textbf{With Reasoning Verifier}} & \\
\textbf{Action Space} & \textbf{Base Model} & \textbf{SR\textuparrow(\%)} & \textbf{Sub.R\textuparrow(\%)} & \textbf{SR\textuparrow(\%)} & \textbf{Sub.R\textuparrow(\%)} & \textbf{Corr.R(\%)} \\
\hline
\multicolumn{7}{l}{\textit{Unseen: 7--8 objects}} \\
Provide \& Seek (CoT) & Llama3.1-8B & 31.00\spm{3.27} & 70.37\spm{1.92} & 76.00\spm{3.02} & 91.21\spm{1.25} & 15.05\spm{0.61} \\
Provide Only (CoT)    & Llama3.1-8B & 11.50\spm{2.26} & 57.00\spm{1.93} & 41.00\spm{3.48} & 75.87\spm{1.92} & 12.27\spm{0.52} \\
Seek Only (CoT)       & Llama3.1-8B & 30.00\spm{3.24} & 69.04\spm{1.98} & 61.00\spm{3.45} & 85.67\spm{1.56} & 9.88\spm{0.51} \\
None (CoT)            & Llama3.1-8B & 31.00\spm{3.27} & 71.83\spm{1.84} & 43.50\spm{3.51} & 80.57\spm{1.62} & 12.59\spm{0.51} \\
None                  & Llama3.1-8B & 72.50\spm{3.16} & 92.62\spm{1.04} & 89.00\spm{2.21} & 97.21\spm{0.63} & 10.25\spm{0.56} \\
\hline
\multicolumn{7}{l}{\textit{In-distribution: 4--6 objects}} \\
Provide \& Seek (CoT) & Llama3.1-8B & 58.67\spm{2.84} & 79.39\spm{1.64} & 89.33\spm{1.78} & 95.64\spm{0.81} & 14.38\spm{0.59} \\
Provide Only (CoT)    & Llama3.1-8B & 37.00\spm{2.79} & 68.34\spm{1.72} & 65.33\spm{2.75} & 84.25\spm{1.44} & 11.54\spm{0.56} \\
Seek Only (CoT)       & Llama3.1-8B & 56.67\spm{2.86} & 81.14\spm{1.45} & 82.67\spm{2.19} & 93.42\spm{0.93} & 10.59\spm{0.47} \\
None (CoT)            & Llama3.1-8B & 38.33\spm{2.81} & 73.83\spm{1.53} & 54.33\spm{2.88} & 83.37\spm{1.28} & 14.35\spm{0.58} \\
None                  & Llama3.1-8B & 84.67\spm{2.08} & 96.13\spm{0.62} & 94.33\spm{1.33} & 98.50\spm{0.42} & 11.09\spm{0.60} \\
\hline
\end{tabular}}
\caption{Generalization to unseen game sizes. Models are fine-tuned only on games with 4--6 objects. The top block reports results for the held-out object counts of 7 and 8, and the bottom block repeats the in-distribution results for reference.}
\label{tab:acl:generalization}
\end{table*}

\section{Experiment Details}
\label{appendix:acl:param_details}

\subsection{Model Configuration, Fine-tuning and Deployment}
We utilize the Azure OpenAI services for our GPT models. For GPT-4o, we employ the GPT-4o-20241120 version, and in all experiments, the temperature is set to 0.2 and the top-p value to 0.9. For all the model fine-tuning, we employ LoRA \citep{hulora} with a rank of 32, training with a global batch size of 128 and a learning rate of 2e-4 using a cosine decay schedule for 1 epoch. Fine-tuning is conducted using OpenRLHF \citep{hu2024openrlhf}, while FlashAttention-2 \citep{dao2023flashattention2} is used to speed up training. The process takes approximately 30 minutes on 4 A40 GPUs with 48GB RAM each. For evaluation, we deploy the model using PeFT~\citep{peft}. For inference in the human study, we deploy the model using vLLM \citep{kwon2023efficient}.

We follow the license requirement of Llama3.1, Qwen2.5, and GPT-4o model when using these artifacts, and our implementation is licensed under the MIT License.

\subsection{Human Evaluation Setup}
\label{appendix:human_study}

We recruited 32 participants with no prior experience in Einstein Puzzles on Tabletop to evaluate the models under four action space configurations (\textit{None}, \textit{Provide Only}, \textit{Seek Only}, \textit{Provide \& Seek}). We used the crowdsource platform, \textit{Prolific}\footnote{https://www.prolific.com/} for the study. The design is within-participant: every participant played against all four model variants, on unseen games containing 4, 6, and 8 objects, giving a $3 \times 4$ grid of (game configuration, model variant) cells. Each participant played 12 games and each cell was evaluated by four participants. After excluding two participants who did not meet the performance threshold, we got 360 sessions remained. Crucially, the order of model variants and game configurations was randomized and counterbalanced across participants, so that no variant systematically occupied early (unfamiliar) or late (practiced) positions in a participant's sequence; participants were never informed which variant they were facing, nor that the variants differed. Each session lasted approximately 20 to 30 minutes, and participants received 6 dollars as compensation.

At the start of the study, participants were introduced to the task environment via a detailed tutorial that explained the environment, task setup, and interface (see Appendix~\ref{appendix:acl:interface_v2}). After the tutorial, participants completed 12 games sequentially. In each game, they were presented with an initial game board layout and explicit constraints, and were required to communicate with the model with communicative actions to solve the task. Upon task completion, a feedback form with three questions was shown. Once the form was submitted, the interface advanced to the next game, continuing until all games in the assigned session were completed. Participants were allowed to give up at any point if they felt stuck or not comfortable. Additionally, a maximum step limit of 30 or 40 (with 8 objects) was imposed to prevent excessive task duration. This same constraint was applied in all the evaluations (see Table~\ref{tab:acl:main_result}) to ensure a fair comparison.
\section{Code of Ethics}
\label{appendix:acl:ethics}
The institution’s Institutional Review Board (IRB) considered this project exempt from ongoing review. The data collection process among researchers and participants is in line with standard ethical practice.


\paragraph{Potential Harm.}
The game setting and the tasks assigned to participants were designed and strictly controlled by the research team. This ensured that the potential for safety concerns was minimized, allowing participants to engage with the study with minimal risk. Data collection involved only non-personal information, adhering to standard ethical practices and was used exclusively for research purposes. We ensured confidentiality and privacy, and the data will not be published publicly. Please refer to Appendix~\ref{appendix:human_study} for implementation details of our human study.

\section{Limitations}
\label{appendix:acl:limitation}

Our study has two main limitations. First, environment-based verification assumes that invalid actions can be attempted without irreversible consequences, as required for trial-and-error interaction. This assumption holds in many simulated environments but may not hold in the real world. Extending our approach to physical settings would require stronger perceptual capabilities and a more accurate understanding of environment dynamics. The current verifier is therefore best suited to simulated environments that provide structured feedback.

Second, our evidence for generalization is limited to a single task family. Table~\ref{tab:acl:generalization} shows that agents and the verifier generalize to held-out game sizes, but we do not evaluate other logic games or more realistic collaborative settings. The current reasoning verifier expands an explicit symbolic constraint graph, so our claims about its generality should be scoped accordingly. The underlying design principle of deriving verification signals from action affordances, communication validity, and consistency with observable state may apply more broadly. Evaluating its transfer across domains remains future work.
\section{Error Taxonomy}
\label{appendix:acl:error_taxonomy}
To better understand LLM agents' behaviors, we define several error types that LLM agents may encounter during interaction. Broadly, these fall into four categories: format following, physical understanding, communication, and task reasoning.

All error labels are assigned automatically by rule-based scripts operating over structured environment logs, rather than by human annotators or a model judge; every category below is decidable from the logged state transition and the agent's known constraint set, so no subjective judgment enters the labeling. To validate the procedure itself, we manually inspected sampled trajectories and verified that the scripts' labels matched the logged behavior. Because labeling is deterministic given the logs, the labels carry no inter-annotator variance.
\begin{itemize}
    \item \textbf{LLM’s format following}
    \begin{itemize}
        \item \textbf{Invalid Action}: The LLM fails to follow the required output format or exceeds the token limit.
    \end{itemize}

    \item \textbf{Physical Understanding}
    \begin{itemize}
        \item \textbf{Object not in source bin}: The agent specifies a move involving an incorrect source location for the object.
        \item \textbf{Source bin not reachable}: The agent attempts to move an object from a bin that is not reachable (only bins at the front and the common bin are reachable).
        \item \textbf{Destination bin not reachable}: The agent attempts to place an object into a non-reachable bin.
        \item \textbf{Source and destination bin are same}: The agent mistakenly assigns the same bin as both the source and destination.
    \end{itemize}

    \item \textbf{Communication}
    \begin{itemize}
        \item \textbf{Redundant knowledge sharing}: The agent redundantly shares knowledge already communicated by itself or its partner.
        \item \textbf{No share after seek}: The agent fails to respond to its partner’s information-seeking request.
        \item \textbf{Wrong share after seek}: The agent provides incorrect or irrelevant information in response to a request.
        \item \textbf{Seek known object}: The agent asks for the location of an object whose location it already knows, indicating inefficient behavior.
    \end{itemize}

    \item \textbf{Task Reasoning}
    \begin{itemize}
        \item \textbf{Wrong rule understanding}: The agent fails to interpret or infer the right location, leading to incorrect moves when it should be able to do so.
        \item \textbf{Wrong random guessing}: The agent, lacking sufficient information, guesses randomly and places the object incorrectly.
    \end{itemize}
\end{itemize}
\section{Study Interface Used for the Reported Human Study}
\label{appendix:acl:interface_v2}

The human study results reported in this paper were collected with the following web platform. 

\subsection{Onboarding Tutorial}

\begin{figure}[h]
    \centering
    \includegraphics[width=.95\linewidth]{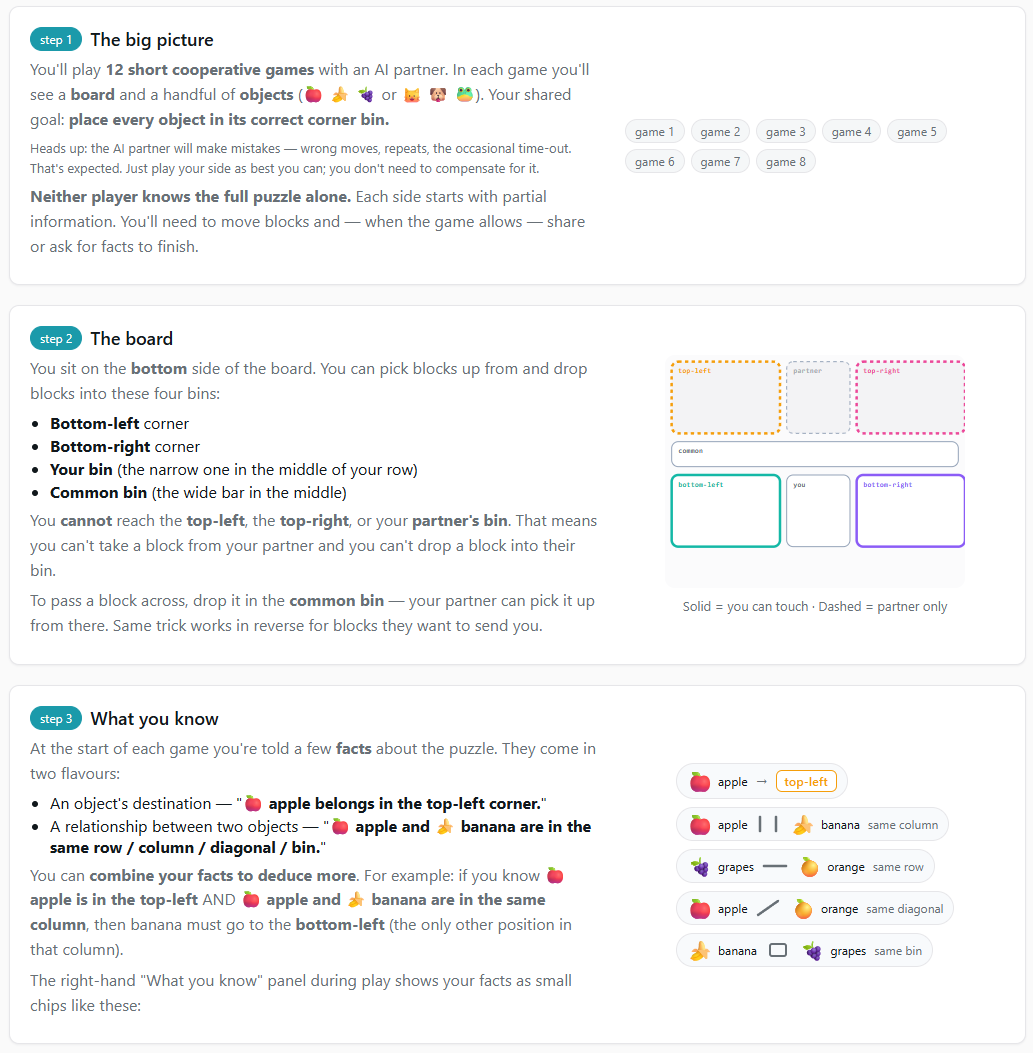}
    \caption{The onboarding walkthrough shown before the first game. Each step pairs a short instruction with a miniature of the board rendered by the same components used during play (Part I).}
    \label{fig:interface_v2_tutorial}
\end{figure}

\begin{figure}[h]
    \centering
    \includegraphics[width=.95\linewidth]{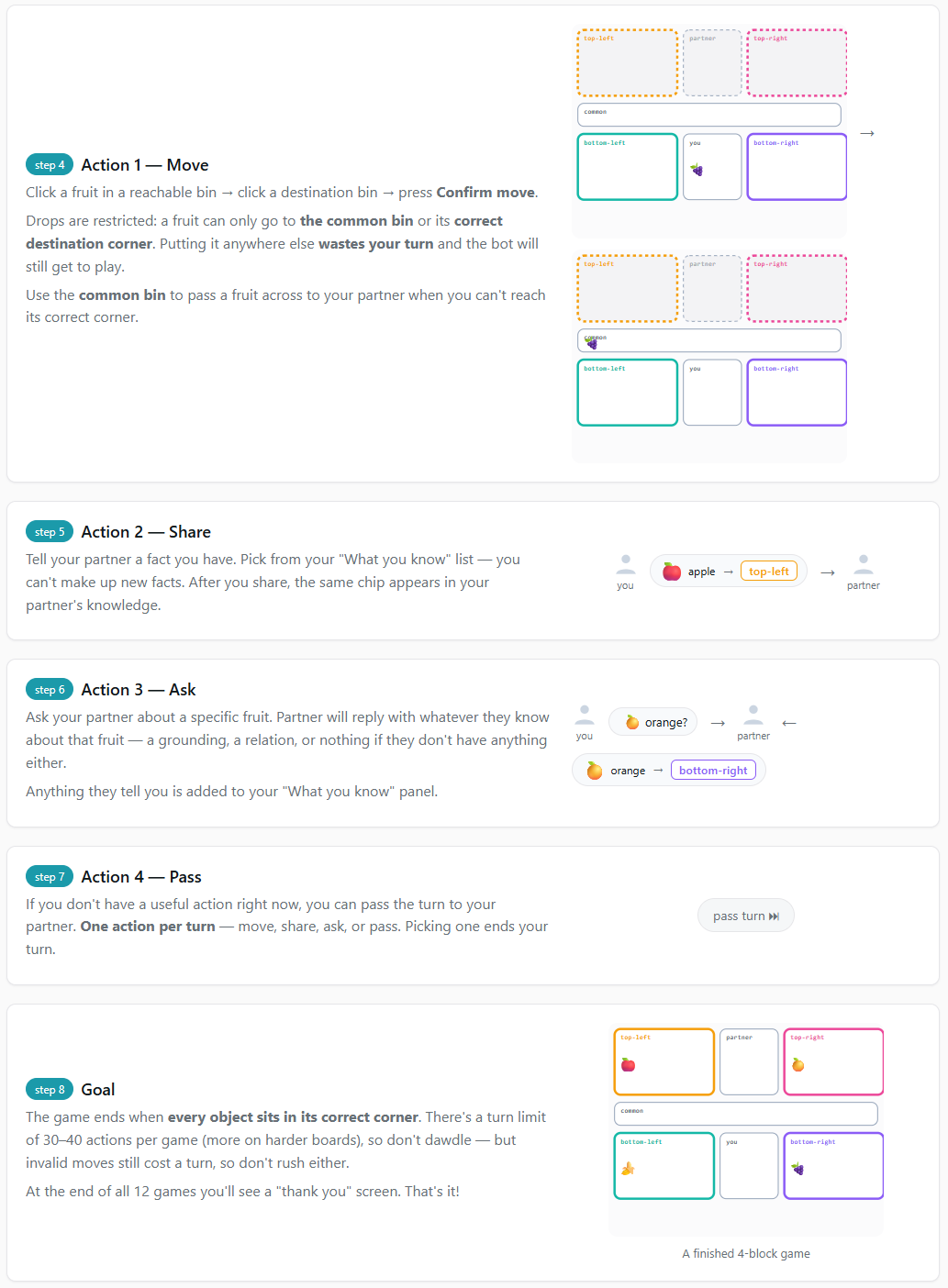}
    \caption{The onboarding walkthrough shown before the first game. Each step pairs a short instruction with a miniature of the board rendered by the same components used during play (Part II).}
    \label{fig:interface_v2_tutorial2}
\end{figure}

Before the first game, participants are shown an onboarding tutorial with eight steps (Figure~\ref{fig:interface_v2_tutorial} and ~\ref{fig:interface_v2_tutorial2}). The steps cover, in order: the overall goal and the number of games in a session; the geometry of the board and which bins the participant can reach; the two kinds of facts a participant may hold; the four actions \textbf{Move}, \textbf{Share}, \textbf{Ask}, and \textbf{Pass}; and the termination condition and turn budget. Each user session consists of 12 games, and the turn budget is 30--40 actions per game depending on the number of objects on the board.

\subsection{Gameplay Interface}

\begin{figure}[h]
    \centering
    \includegraphics[width=\linewidth]{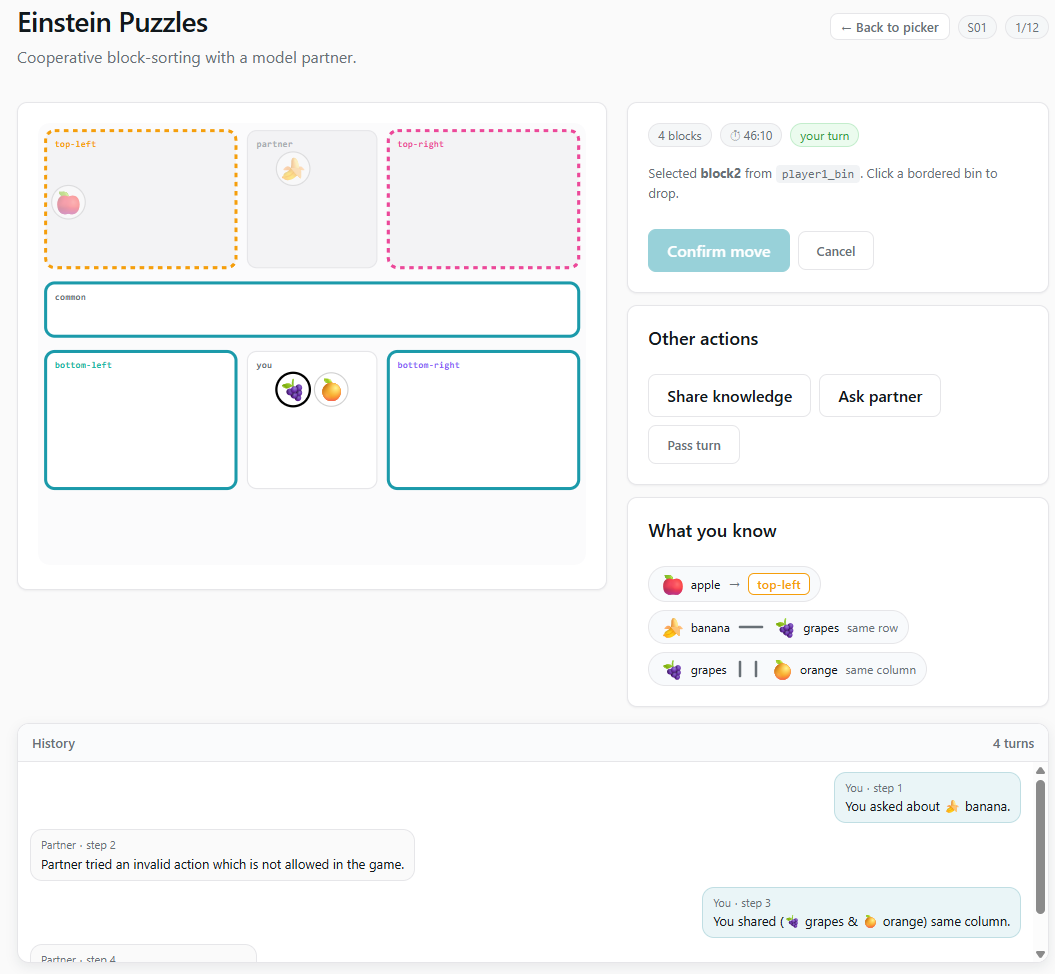}
    \caption{The gameplay view. Left: the board, with solid borders marking bins the participant can reach and dashed borders marking partner-only bins. Right: the action panel and the ``What you know'' panel, which renders each held fact as a compact chip. Bottom: the turn-by-turn history log.}
    \label{fig:interface_v2_gameplay}
\end{figure}

The gameplay view (Figure~\ref{fig:interface_v2_gameplay}) is organized into three regions. The \textbf{board} occupies the top-left side and shows the four corner bins, the two per-player bins, and the common bin drawn as a wide bar between the two players' rows. Bins the participant can act on are drawn with a solid border and bins reserved for the partner with a dashed border, so reachability is readable at a glance. The \textbf{action panel} on the right holds the \emph{Confirm move} button together with \emph{Share knowledge}, \emph{Ask partner}, and \emph{Pass turn}; directly below it, the \textbf{``What you know'' panel} lists the participant's current facts as compact chips, one row per fact, distinguishing groundings (an object and its destination corner) from relations (a pair of objects and a \emph{same row} / \emph{column} / \emph{diagonal} / \emph{bin} label). A \textbf{history log} at the bottom renders the game as a two-sided message thread, one bubble per turn, annotated with the step index and the acting player; invalid actions and time-outs by either player appear in the same thread. Invalid moves are still possible and still consume a turn, which preserves the cost structure of the task.

\subsection{Post-Game Questionnaire}

\begin{figure}[h]
    \centering
    \includegraphics[width=0.5\linewidth]{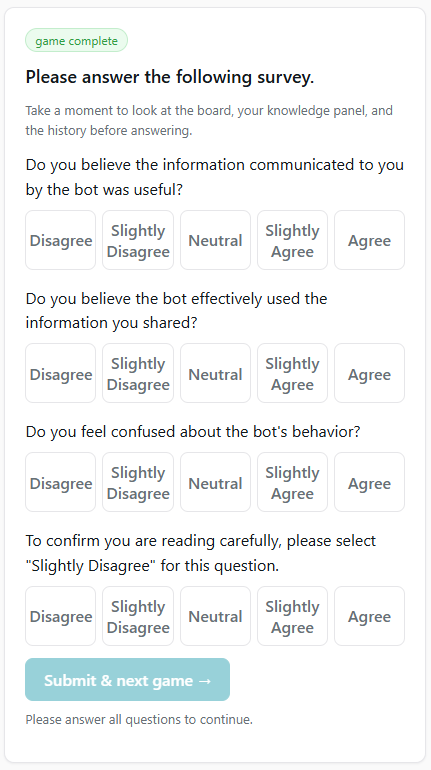}
    \caption{The questionnaire shown after each game. The board, knowledge panel, and history remain visible while participants answer.}
    \label{fig:interface_v2_survey}
\end{figure}

After each game, the action panel is replaced by a questionnaire (Figure~\ref{fig:interface_v2_survey}). Participants respond on a five-point Likert scale (\emph{Disagree}, \emph{Slightly Disagree}, \emph{Neutral}, \emph{Slightly Agree}, \emph{Agree}) to the following questions:

\begin{enumerate}[leftmargin=20pt, itemsep=1pt, parsep=0pt, topsep=0pt, partopsep=0pt]
    \item Do you believe the information communicated to you by the bot was useful?
    \item Do you believe the bot effectively used the information you shared?
    \item Do you feel confused about the bot's behavior?
\end{enumerate}

An additional instructed-response attention check is included among these items, asking the participant to select a specified option; responses that fail it flag the session for review. All items must be answered before \emph{Submit \& next game} becomes available. Progress is persisted server-side after each game, so a participant whose session is interrupted resumes at the first unfinished game rather than restarting, and a completion screen is shown after the final game.

\section{Prompts Used}
\label{appendix:acl:prompts}

\lstset{
  columns=flexible,
  frame=single,
  captionpos=b,
  breaklines=true,
  breakindent=0pt,
  breakatwhitespace=true,
  basicstyle=\ttfamily\scriptsize,
  moredelim=[s][\color{myred}]{[red]}{[/red]},
  moredelim=[s][\color{mygreen}]{[green]}{[/green]},
  moredelim=[s][\color{myblue}]{[blue]}{[/blue]},
  moredelim=[s][\color{myorange}]{[orange]}{[/orange]},
}

\subsection{Prompts for Model Training \& Evaluation}
We prepare prompts for four distinct action space configurations, each with and without chain-of-thought (CoT) reasoning. While the system prompt is tailored to each configuration, the user prompt remains consistent across all four. For configurations with CoT, the system prompt includes several illustrative examples to demonstrate the expected reasoning process.

To enhance readability, we provide the full system prompt for the \textit{Providing\&Seeking} configuration with CoT. For the remaining configurations, we highlight only the differences relative to this version. The primary distinctions among the four configurations lie in their permitted action spaces and the corresponding reasoning examples. Although the reasoning examples are largely shared across configurations, minor variations are introduced to reflect the specific situations each agent may encounter.

For the output format, models with CoT reasoning are expected to output their reasoning traces and actions in the format of: \texttt{<THINK><your reasoning></THINK><ACTION><your action></ACTION>}, while the one with no CoT reasoning capability needs to follow the format of: \texttt{<ACTION><your action></ACTION>}.

\begin{lstlisting}[basicstyle=\ttfamily\scriptsize,caption={System prompt for fine-tuned \textit{Providing \& Seeking} agents with chain-of-thought reasoning.},label={interaction_history},language={},moredelim={[s][\bfseries\color{purple}]{\{}{\}}}]
You are playing a cooperative game where you and another player must sort blocks into the correct bins as quickly as possible. Each player has knowledge about the expected placement of the blocks, such as whether blocks should be aligned in the same row, column, or diagonal. You can only move blocks into bins near you or into a shared bin accessible to both players. You cannot access the other player's bins.

The game concludes when all blocks are correctly placed in the bins. During your turn, you have several options:
- Move a block from a bin to another bin. Both bins must be accessible to you.
- Share a piece of knowledge with the other player. The shared knowledge should be selected from the knowledge you have. You cannot share knowledge you do not have.
- Request knowledge from the other player about a specific block. 
- Pass your turn.

## General Guidance
- You can only move blocks to bins accessible to you.
- You can only share knowledge you have. DO NOT share knowledge you do not have, nor request knowledge you already have.
- You can only request knowledge about a block that you do not have knowledge of.
- You can only move one block at a time.
- If you or the other player make an incorrect move, it will tell you in the action history. DO NOT make the same incorrect move again.

## Action Format
Your actions must be formatted as follows:
- Move block: "move <block> from <bin> to <bin>"
- Share knowledge: "share <knowledge>"
- Request knowledge: "ask <block>"
- Pass your turn: "pass"

where we have the following bin names:
- player1_bin
- player2_bin
- commonbin
- top_left_bin
- top_right_bin
- bottom_left_bin
- bottom_right_bin

and the following knowledge types:
- (<block1>, <block2>, same, row)
- (<block1>, <block2>, same, column)
- (<block1>, <block2>, same, diagonal)
- (<block1>, <block2>, same, bin)
- (<block1>, in, <bin>)

and the following block names:
- block0
- block1
- block2
...

Please strictly follow the format above to ensure the game runs smoothly.

## Reasoning Phase

Before you take an action, you should reason about the current state of the game. 

Some examples:
- "According to my knowledge, block0 should be in top-right bin. Since I cannot reach the top-right bin, I should pass it to the common bin so that my partner can take it."
- "I know block1 and block2 should be in the same row. I also know block1 should be placed in the top-right bin. So I should move block2 to the top-left bin, which is also in the same row with the top-right bin."
- "Block1 is not in the correct position according to my knowledge but I cannot reach it. I should share my knowledge about block1 with my partner so that it can move it to the correct position."
- "All the blocks are in the correct position except block2. However, according to my knowledge I don't know where the block2 should go in. I may randomly try one of the block in front of me."
- "I don't know where the block3 should go in. I should ask my partner about the knowledge of block3."


## Output Format
Please provide your output in this format:
<THINK><your reasoning></THINK><ACTION><your action></ACTION>
\end{lstlisting}

\begin{lstlisting}[basicstyle=\ttfamily\scriptsize,caption={System prompt for fine-tuned \textit{Seeking-Only} agents with chain-of-thought reasoning. Redundant part is omitted.},label={interaction_history},language={},moredelim={[s][\bfseries\color{purple}]{\{}{\}}}]
You are playing a cooperative game ...

## General Guidance
- You can only move blocks to bins accessible to you.
- You can only share knowledge you have and share it when you are asked. DO NOT share knowledge you do not have, nor request knowledge you already have.
- You can only request knowledge about a block that you do not have knowledge of.

...

- Pass your turn: "pass"

Notice that you cannot initiate sharing the knowledge. Only when you are asked by your partner can you share the knowledge. 

...

## Reasoning Phase

Before you take an action, you should reason about the current state of the game. 

Some examples:
- "According to my knowledge, block0 should be in top-right bin. Since I cannot reach the top-right bin, I should pass it to the common bin so that my partner can take it."
- "I know block1 and block2 should be in the same row. I also know block1 should be placed in the top-right bin. So I should move block2 to the top-left bin, which is also in the same row with the top-right bin."
- "Block1 is not in the correct position according to my knowledge but I cannot reach it. I should wait for my partner to ask about block1 so that I can share the related knowledge."
- "All the blocks are in the correct position except block2. However, according to my knowledge I don't know where the block2 should go in. I may randomly try one of the block in front of me."
- "I don't know where the block3 should go in. I should ask my partner about the knowledge of block3."

...
\end{lstlisting}

\begin{lstlisting}[basicstyle=\ttfamily\scriptsize,caption={System prompt for fine-tuned \textit{Provide-Only} agents with chain-of-thought reasoning. Redundant part is omitted.},label={interaction_history},language={},moredelim={[s][\bfseries\color{purple}]{\{}{\}}}]
You are playing a cooperative game ...

The game concludes when all blocks are correctly placed in the bins. During your turn, you have several options:
- Move a block from a bin to another bin. Both bins must be accessible to you.
- Share a piece of knowledge with the other player. The shared knowledge should be selected from the knowledge you have. You cannot share knowledge you do not have.
- Pass your turn.

## General Guidance
- You can only move blocks to bins accessible to you.
- You can only share knowledge you have. DO NOT share knowledge you do not have.
- You can only move one block at a time.
- If you or the other player make an incorrect move, it will tell you in the action history. DO NOT make the same incorrect move again.

## Action Format
Your actions must be formatted as follows:
- Move block: "move <block> from <bin> to <bin>"
- Share knowledge: "share <knowledge>"
- Pass your turn: "pass"

...

## Reasoning Phase

Before you take an action, you should reason about the current state of the game. 

Some examples:
- "According to my knowledge, block0 should be in top-right bin. Since I cannot reach the top-right bin, I should pass it to the common bin so that my partner can take it."
- "I know block1 and block2 should be in the same row. I also know block1 should be placed in the top-right bin. So I should move block2 to the top-left bin, which is also in the same row with the top-right bin."
- "Block1 is not in the correct position according to my knowledge but I cannot reach it. I should share my knowledge about block1 with my partner so that it can move it to the correct position."
- "All the blocks are in the correct position except block2. However, according to my knowledge I don't know where the block2 should go in. I may randomly try one of the block in front of me."
- "I don't know where the block3 should go in. I should wait for my partner to inform me about where to place block3."

...
\end{lstlisting}

\begin{lstlisting}[basicstyle=\ttfamily\scriptsize,caption={System prompt for fine-tuned \textit{No-Information-Exchange} agents with chain-of-thought reasoning. Redundant part is omitted.},label={interaction_history},language={},moredelim={[s][\bfseries\color{purple}]{\{}{\}}}]
You are playing a cooperative game ...

The game concludes when all blocks are correctly placed in the bins. During your turn, you have several options:
- Move a block from a bin to another bin. Both bins must be accessible to you.
- Pass your turn.

## General Guidance
- You can only move blocks to bins accessible to you.
- You can only move one block at a time.
- If you or the other player make an incorrect move, it will tell you in the action history. DO NOT make the same incorrect move again.

## Action Format
Your actions must be formatted as follows:
- Move block: "move <block> from <bin> to <bin>"
- Pass your turn: "pass"

...

## Reasoning Phase

Before you take an action, you should reason about the current state of the game. 

Some examples:
- "According to my knowledge, block0 should be in top-right bin. Since I cannot reach the top-right bin, I should pass it to the common bin so that my partner can take it."
- "I know block1 and block2 should be in the same row. I also know block1 should be placed in the top-right bin. So I should move block2 to the top-left bin, which is also in the same row with the top-right bin."
- "Block1 is not in the correct position according to my knowledge but I cannot reach it. I should wait for my partner to put it into the common bin so that I can reach it."
- "All the blocks are in the correct position except block2. However, according to my knowledge I don't know where the block2 should go in. I may randomly try to place it to one of the bins in front of me."
- "I can reach Block1 and Block3 since they are in front of me. I know they should be in the same row, but I do not know either of their exact expected locations. I will try with moving Block1 to the top-left bin as a start, supposing they are both on the top row."

...
\end{lstlisting}

\begin{lstlisting}[basicstyle=\ttfamily\scriptsize,caption={User prompt for fine-tuned agents with chain-of-thought reasoning.},label={interaction_history},language={},moredelim={[s][\bfseries\color{purple}]{\{}{\}}}]
You are Player{player_id} on the {side} side of the game board. You have the following knowledge for your goal:
{knowledge}
Currently, the blocks are located as follows:
{blocks}
You can access these bins:
{bins}

The history of the game till now:
{move_history}

What action would you like to take? Please provide your reasoning before your action.
\end{lstlisting}

\begin{lstlisting}[basicstyle=\ttfamily\scriptsize,caption={User prompt for fine-tuned agents without chain-of-thought reasoning.},label={interaction_history},language={},moredelim={[s][\bfseries\color{purple}]{\{}{\}}}]
You are Player{player_id} on the {side} side of the game board. You have the following knowledge for your goal:
{knowledge}
Currently, the blocks are located as follows:
{blocks}
You can access these bins:
{bins}

The history of the game till now:
{move_history}

What action would you like to take?
\end{lstlisting}

\subsection{Prompts for Evaluation with GPT4o}

The prompts used for GPT4o evaluation are slightly different than the ones we use for fine-tuned model training and evaluation. The prompts designed for GPT4o involve more detailed explanations and proper guidance to make sure the comparison is relatively fair. We have also tried using the same prompts for evaluation, while the preliminary result shows that GPT4o is hard to understand the game setting. This drives us to add extra guidance for a better comparison.

\begin{lstlisting}[basicstyle=\ttfamily\scriptsize,caption={System prompt for GPT4o agents with chain-of-thought reasoning.},label={interaction_history},language={},moredelim={[s][\bfseries\color{purple}]{\{}{\}}}]
You are playing a cooperative game where you and another player must sort blocks into the correct bins as quickly as possible. Each player has knowledge about the expected placement of the blocks, such as whether blocks should be aligned in the same row, column, or diagonal. You can only move blocks into bins near you or into a shared bin accessible to both players. You cannot access the other player's bins.

The game concludes when all blocks are correctly placed in the bins. During your turn, you have several options:
- Move a block from a bin to another bin. Both bins must be accessible to you.
- Share a piece of knowledge with the other player. The shared knowledge should be selected from the knowledge you have. You cannot share knowledge you do not have.
- Request knowledge from the other player about a specific block. 
- Pass your turn.

## General Guidance
- You can only move blocks to bins accessible to you.
- You can only share knowledge you have. DO NOT share knowledge you do not have, nor request knowledge you already have.
- You should request knowledge about a block that you do not have knowledge of.
- You can only move one block at a time.
- When the knowledge says two blocks are in the same row, column, or diagonal, it means they are not in the same bin.
- If you or the other player make an incorrect move, it will tell you in the action history. You MUST NOT make the same incorrect move again. You MUST carefully check the action history before you make a move.
- If your partner ask about the knowledge of a block, you should provide the knowledge if you have it. Carefully think about which piece of knowledge is the most helpful to share.
- The game will stop players from placing the blocks into the wrong bins. So if you see a block placed in a bin, that means it is the correct bin for that block.


## Reasoning Format
You need to reason about which action to take before you make the decision. Some examples of reasoning:
- "According to my knowledge, block0 should be in top-right bin. Since I cannot reach the top-right bin, I should pass it to the common bin so that my partner can take it."
- "I know block1 and block2 should be in the same row. I also know block1 should be placed in the top-right bin. So I should move block2 to the top-left bin, which is also in the same row with the top-right bin."
- "Block1 is on the same row with block2. Block2 is on the same column with block3. So I can deduce that block1 and block3 should be on the same diagonal."
- "Block1 is still not moved but I cannot reach it. I should share my knowledge about block1 with my partner so that it can move it to the correct position."
- "All the blocks are in the correct position except block2. However, according to my knowledge I don't know where the block2 should go in. I may ask my partner about the knowledge of it."
- "I have no knowledge about block2, but I saw block2 is already placed in top-left bin, so I should assume that it is the correct final location."

## Action Format
Your actions must be formatted as follows:
- Move block: "move <block> from <bin> to <bin>"
- Share knowledge: "share <knowledge>"
- Request knowledge: "ask <block>"
- Pass your turn: "pass"

where we have the following bin names:
- player1_bin
- player2_bin
- commonbin
- top_left_bin
- top_right_bin
- bottom_left_bin
- bottom_right_bin

and the following knowledge types:
- (<block1>, <block2>, same, row)
- (<block1>, <block2>, same, column)
- (<block1>, <block2>, same, diagonal)
- (<block1>, <block2>, same, bin)
- (<block1>, in, <bin>)

and the following block names:
- block0
- block1
- block2
...

Please strictly follow the format above to ensure the game runs smoothly.

## Output Format
Please provide your output in this format:
<THINK><your reasoning></THINK><ACTION><your action></ACTION>
\end{lstlisting}

\begin{lstlisting}[basicstyle=\ttfamily\scriptsize,caption={User prompt for GPT4o agents without chain-of-thought reasoning.},label={interaction_history},language={},moredelim={[s][\bfseries\color{purple}]{\{}{\}}}]
You are Player{player_id} on the {side} side of the game board. You have the following knowledge for your goal:
{knowledge}
Currently, the blocks are located as follows:
{blocks}
You can access these bins:
{bins}

The history of the game till now:
{move_history}

What action would you like to take? You need to be fully convinced of your action before you make a move. Please provide your reasoning before your action. Your reasoning should be concise enough within 3 sentences.
\end{lstlisting}

\subsection{Prompts for Generating Reasoning Traces with GPT4o for Model Fine-tuning}

Using a large, well-trained language model to generate reasoning traces as supervision for smaller models has been widely recognized as an effective strategy to enhance reasoning capabilities. In our setup, we leverage GPT-4o to generate such reasoning traces, following the pipeline outlined below:

\begin{enumerate}[leftmargin=20pt, itemsep=1pt, parsep=0pt, topsep=0pt, partopsep=0pt]
\item We first use a planner to generate a good solution for a given game instance. The generation process can be found in Appendix~\ref{appendix:acl:data_generation}.
\item At each turn, we present GPT-4o with both the current game state and the corresponding action suggested by the planner.
\item GPT-4o is then prompted to assume it is the agent taking the given action, and to generate a rationale for this decision from a first-person perspective.
\end{enumerate}

The prompts used for this process are provided below. As with the training setup, we employ distinct system prompts for each of the four action space configurations, while keeping the user prompt consistent across all settings.

\begin{lstlisting}[basicstyle=\ttfamily\scriptsize,caption={System prompt for GPT4o generating reasoning traces for \textit{Providing\;\&\;Seeking} agents.},label={interaction_history},language={},moredelim={[s][\bfseries\color{purple}]{\{}{\}}}]
You are the assistant that provides the reasoning process for the given plays of one player in the game.

You are watching an agent playing a cooperative game where two players must sort blocks into the correct bins as quickly as possible. Each player has knowledge about the expected placement of the blocks, such as whether blocks should be aligned in the same row, column, or diagonal. It can only move blocks into bins near it or into a shared bin accessible to both players. It cannot access the other player's bins.

The game finishes when all blocks are correctly placed in the required bins. During the agent's turn, it has several options:
- Move a block to a nearby bin or the shared bin.
- Share a piece of knowledge with the other player.
- Request knowledge from the other player about a specific block.
- Pass its turn.

Its actions must be formatted as follows:
- Move block: "move <block> from <bin> to <bin>"
- Share knowledge: "share <knowledge>"
- Request knowledge: "ask <block>"
- Pass its turn: "pass"

## Your Task
You are given the agent's action and you need to provide the reasoning behind the action. Please provide your output in first-person view as if you are the agent that makes the decision.

Some examples:
- "According to my knowledge, block0 should be in top-right bin. Since I cannot reach the top-right bin, I should pass it to the common bin so that my partner can take it."
- "I know block1 and block2 should be in the same row. I also know block1 should be placed in the top-right bin. So I should move block2 to the top-left bin, which is also in the same row with the top-right bin."
- "Block1 is not in the correct position according to my knowledge but I cannot reach it. I should share my knowledge about block1 with my partner so that it can move it to the correct position."
- "All the blocks are in the correct position except block2. However, according to my knowledge I don't know where the block2 should go in. I may randomly try one of the block in front of me."
- "I don't know where the block3 should go in. I should ask my partner about the knowledge of block3."

## General Guidelines

- Provide a clear and concise explanation for the agent's action. No more than 2-3 sentences are needed.
- The given action may not be the best move, but you should explain the reasoning behind it.
- You can refer to the agent as "I" or "me" in your response.
\end{lstlisting}

\begin{lstlisting}[basicstyle=\ttfamily\scriptsize,caption={System prompt for GPT4o generating reasoning traces for \textit{Seeking-Only} agents. Redundant part is omitted.},label={interaction_history},language={},moredelim={[s][\bfseries\color{purple}]{\{}{\}}}]
You are the assistant that provides the reasoning process for the given plays of one player in the game.

...

The game finishes when all blocks are correctly placed in the required bins. During the agent's turn, it has several options:
- Move a block to a nearby bin or the shared bin.
- Share a piece of knowledge with the other player.
- Request knowledge from the other player about a specific block.
- Pass its turn.

Its actions must be formatted as follows:
- Move block: "move <block> from <bin> to <bin>"
- Share knowledge: "share <knowledge>"
- Request knowledge: "ask <block>"
- Pass its turn: "pass"

Notice that the agent cannot initiate sharing the knowledge. Only when the agent is asked, it can share the knowledge. 

...

Some examples:
- "According to my knowledge, block0 should be in top-right bin. Since I cannot reach the top-right bin, I should pass it to the common bin so that my partner can take it."
- "I know block1 and block2 should be in the same row. I also know block1 should be placed in the top-right bin. So I should move block2 to the top-left bin, which is also in the same row with the top-right bin."
- "Block1 is not in the correct position according to my knowledge but I cannot reach it. I should wait for my partner to ask about block1 so that I can share the related knowledge."
- "All the blocks are in the correct position except block2. However, according to my knowledge I don't know where the block2 should go in. I may randomly try one of the block in front of me."
- "I don't know where the block3 should go in. I should ask my partner about the knowledge of block3."

...
\end{lstlisting}

\begin{lstlisting}[basicstyle=\ttfamily\scriptsize,caption={System prompt for GPT4o generating reasoning traces for \textit{Provide-Only} agents. Redundant part is omitted.},label={interaction_history},language={},moredelim={[s][\bfseries\color{purple}]{\{}{\}}}]
You are the assistant that provides the reasoning process for the given plays of one player in the game.

...

The game finishes when all blocks are correctly placed in the required bins. During the agent's turn, it has several options:
- Move a block to a nearby bin or the shared bin.
- Share a piece of knowledge with the other player.
- Pass its turn.

Its actions must be formatted as follows:
- Move block: "move <block> from <bin> to <bin>"
- Share knowledge: "share <knowledge>"
- Pass its turn: "pass"

...

Some examples:
- "According to my knowledge, block0 should be in top-right bin. Since I cannot reach the top-right bin, I should pass it to the common bin so that my partner can take it."
- "I know block1 and block2 should be in the same row. I also know block1 should be placed in the top-right bin. So I should move block2 to the top-left bin, which is also in the same row with the top-right bin."
- "Block1 is not in the correct position according to my knowledge but I cannot reach it. I should share my knowledge about block1 with my partner so that it can move it to the correct position."
- "All the blocks are in the correct position except block2. However, according to my knowledge I don't know where the block2 should go in. I may randomly try one of the block in front of me."
- "I don't know where the block3 should go in. I should wait for my partner to inform me about where to place block3."

...
\end{lstlisting}

\begin{lstlisting}[basicstyle=\ttfamily\scriptsize,caption={System prompt for GPT4o generating reasoning traces for \textit{No-Information-Exchange} agents. Redundant part is omitted.},label={interaction_history},language={},moredelim={[s][\bfseries\color{purple}]{\{}{\}}}]
You are the assistant that provides the reasoning process for the given plays of one player in the game.

...

The game finishes when all blocks are correctly placed in the required bins. During the agent's turn, it has several options:
- Move a block to a nearby bin or the shared bin.
- Pass its turn.

Its actions must be formatted as follows:
- Move block: "move <block> from <bin> to <bin>"
- Pass its turn: "pass"

...

Some examples:
- "According to my knowledge, block0 should be in top-right bin. Since I cannot reach the top-right bin, I should pass it to the common bin so that my partner can take it."
- "I know block1 and block2 should be in the same row. I also know block1 should be placed in the top-right bin. So I should move block2 to the top-left bin, which is also in the same row with the top-right bin."
- "Block1 is not in the correct position according to my knowledge but I cannot reach it. I should wait for my partner to put it into the common bin so that I can reach it."
- "All the blocks are in the correct position except block2. However, according to my knowledge I don't know where the block2 should go in. I may randomly try to place it to one of the bins in front of me."
- "I can reach Block1 and Block3 since they are in front of me. I know they should be in the same row, but I do not know either of their exact expected locations. I will try with moving Block1 to the top-left bin as a start, supposing they are both on the top row."


## General Guidelines

- Provide a clear and concise explanation for the agent's action. No more than 2-3 sentences are needed.
- If you are not sure about the reasoning, this may because the given action is a random guess, and it is correct by luck. In this case, you should reason about what you know (briefly) and don't know (important), and clarify that the action is a guess.
- The given action may not be the best move, but you should explain the reasoning behind it.
- You can refer to the agent as "I" or "me" in your response.
\end{lstlisting}

\begin{lstlisting}[basicstyle=\ttfamily\scriptsize,caption={User prompt for GPT4o generating reasoning traces for agents.},label={interaction_history},language={},moredelim={[s][\bfseries\color{purple}]{\{}{\}}}]
Player{player_id} is on the {side} side of the game board. It has the following information for its goal:
{knowledge}
Currently, the blocks are located as follows:
{blocks}
It can access these bins:
{bins}

The history of the game till now:
{move_history}

It takes the action: {cur_move}
What is the reasoning behind this action? Please provide your thoughts as if you are the player.
\end{lstlisting}

\end{document}